\documentclass[lettersize,journal]{IEEEtran}
\usepackage{times}

\usepackage[numbers]{natbib}
\usepackage{multicol}
\usepackage[bookmarks=true]{hyperref}

\usepackage{graphics} 
\usepackage{epsfig} 
\usepackage{times} 
\usepackage{amsmath} 
\usepackage{amssymb}  
\usepackage{color}
\usepackage{bm}
\usepackage{subcaption}
\usepackage{booktabs}
\usepackage{enumerate}
\usepackage{todonotes}

\usepackage{textcomp}

\newcommand{\etal}{\textit{et al.~}}

\usepackage[linesnumbered,ruled,lined]{algorithm2e}
\IEEEoverridecommandlockouts

\begin{document}
\title{Enhancing Dexterity in Confined Spaces:\\Real-Time Motion Planning for\\Multi-Fingered In-Hand Manipulation}

\author{Xiao Gao$^{*}$, Kunpeng Yao, Farshad~Khadivar, and Aude~Billard%
\thanks{* Corresponding author: xiao.gao@epfl.ch}
\thanks{All authors are with the Learning Algorithms and Systems Laboratory (LASA), Swiss Federal Institute of Technology in Lausanne (EPFL), Switzerland.}
\thanks{This work was supported by the European Research Council (ERC) through the Advanced Grant No. 741945, Skill Acquisition in Humans and Robots (SAHR).}
}

\maketitle

\begin{abstract}
Dexterous in-hand manipulation in robotics, particularly with multi-fingered robotic hands, poses significant challenges due to the intricate avoidance of collisions among fingers and the object being manipulated.
Collision-free paths for all fingers must be generated in real-time, 
as the rapid changes in hand and finger positions necessitate instantaneous recalculations to prevent collisions and ensure undisturbed movement.
This study introduces a real-time approach to motion planning in high-dimensional spaces.
We first explicitly model the collision-free space using neural networks that are retrievable in real time.
Then, we combined the C-space representation with closed-loop control via dynamical system and sampling-based planning approaches.
This integration enhances the efficiency and feasibility of path-finding, enabling dynamic obstacle avoidance, thereby advancing the capabilities of multi-fingered robotic hands for in-hand manipulation tasks.
\end{abstract}


\section{Introduction}\label{sec::intro}
In the field of robotics, in-hand manipulation is crucial for expanding robotic task capabilities and achieving human-like dexterity in robotic systems. This complex problem, especially with multi-fingered robotic hands, requires utilizing the hand's numerous degrees of freedom to achieve delicate interactions with objects through planned sequences of finger movements.

\begin{figure}[t]
    \centering
    \includegraphics[width=0.48\textwidth]{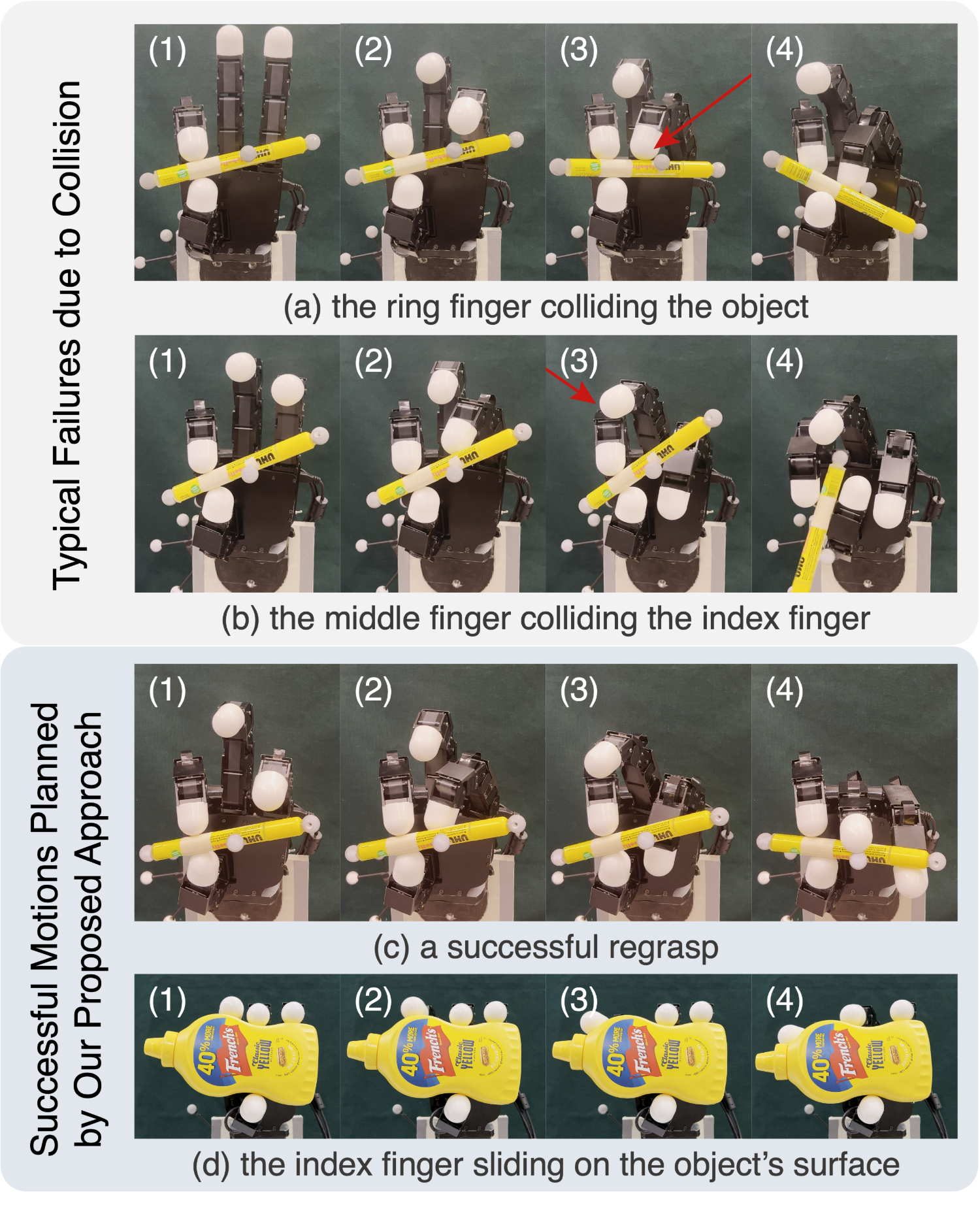}
    \caption{Examples of challenging in-hand manipulation motions demonstrated on a 16-DoF Allegro robotic hand consisting of four fingers (4-DoF each) and a palm. Failures often happen due to (a) finger-object collision or (b) finger-finger collision. Our proposed algorithm enables (c) successful regrasp and (d) finger sliding motion, by generating collision-free motions of multiple fingers in real time.
    \label{fig:regrasp_example}}
\end{figure}

Manipulating an object in hand often involves a sequence of finger motions that transition the object from one {\em grasp} to another until the desired final pose is achieved. To achieve the desired dynamics of the object, the hand can either slide the object alongside the fingers \cite{paolini2014data}, rotate the object among fingers \cite{brock1988enhancing}, or perform a combination of those motions \cite{wan2019regrasp}. Despite significant progress in dexterous robotic control, generating collision-free in-hand manipulation trajectories in real-time remains a challenging task. The complexity lies in navigating the high-dimensional landscape of potential finger movements while ensuring object stability and avoiding unintended finger-object and finger-finger contacts in real-time (see Fig.~\ref{fig:regrasp_example}).
This lack of capability highlights the necessity for innovative methods to accurately model a robotic hand's collision-free space and utilize this model for efficient and effective motion planning.

Addressing this need, this work presents a real-time motion planning framework that models the hand's collision-free space using a set of learned distance functions. These functions are readily accessible in real-time and are integrated with a dynamical systems control paradigm and sampling-based planning techniques. This approach accelerates the search for viable manipulation pathways. This combination of methods not only enables the quick creation of in-hand manipulation techniques, including advanced sliding movements, but also establishes a new standard for efficiency and effectiveness in controlling robotic hands.

Our contributions are threefold.
Firstly, we present a generalized implicit representation of the configuration space (C-space) that is tailored for robotic motion planning and accounts for both self-collisions and hand-object collisions.
Secondly, we introduce a fast and reactive motion planning algorithm that is capable of real-time operation in dynamic and cluttered settings.
Finally, we pioneer a novel technique for in-hand sliding manipulation based on our collision-free motion planning framework.
To the best of our knowledge, this is the first investigation into real-time collision avoidance and sliding planning for a multi-fingered robotic hand.
The validation of our methods on a real multi-fingered robotic hand demonstrates online adaptation to prevent collisions and facilitate object sliding, highlighting its potential to significantly enhance robotic in-hand manipulation capabilities.
\section{Background}\label{sec::related}

\subsection{Collision-free finger motion generation}
Collision detection is a fundamental aspect of collision-free motion planning, involving the computation of distances between robots and obstacles.
The planning of collision-free finger motions for in-hand manipulation can be either model-based or model-free.

Model-based approaches often demand explicit models of the robotic hand and the manipulated object, such as geometric shapes, mass, inertia, and contact models.
For example, Rohrdanz \etal\cite{rohrdanz1997generating} generated regrasp motion by searching in the space of compatible regrasp operations, considering the constraints and grasp quality.
Xue \etal\cite{xue2008planning} planned regrasp motion by constructing a collision-free grasp database offline, then online searching considering collision and kinematic feasibility.
In both tasks, the object is pinched by fingertips, and the trajectories of both fingers and object are assumed collision-free.

Alternatively, optimization-based approaches can naturally ensure collision-free property by incorporating them as problem constraints.
These approaches have been successfully applied to generate grasps \cite{el2013generation} or manipulation motions \cite{sundaralingam2018geometric}.
However, purely model-based approaches are sensitive to model inaccuracies and can hardly capture dynamically changing environments.
In contrast, model-free manipulation approaches usually do not require explicit models of the hand nor of the object, and, instead, use a \emph{policy} to determine the best \emph{action} at each \emph{state} (e.g., configuration).
Andrychowicz \etal\cite{andrychowicz2020learning} combines reinforcement learning with a deep neural network to control a dexterous robotic hand to manipulate a Rubik's cube.
Nagabandi \etal\cite{nagabandi2020deep} trained a deep neural network to enable a dexterous robotic hand to perform a variety of multiple dexterous manipulation motions.
These approaches, however, require a long training time with large amounts of data.
The policy needs to be retrained when manipulating novel objects.

\subsection{Trajectory planning in dynamic environment}
Generating a collision-free trajectory for a finger in real-time becomes more challenging in a dynamic environment, for example, when the manipulated object or other fingers are also moving.
Many works tried to tackle this challenge using an optimal control-based approach.
For example, iterative Linear Quadratic Regulator (iLQR) \cite{todorov2005generalized} used for locally-optimal feedback control of nonlinear dynamical systems; or CHOMP \cite{ratliff2009chomp} that minimizes the cost function by gradient descent.
However, such approaches are prone to be trapped in local minima.
Alternatively, STOMP \cite{Kalakrishnan_STOMP_2011} considers the probabilities of feasible paths and exploits expectation-maximization to obtain a trajectory for the next step.
Both CHOMP and STOMP use a spherical approximation of the robot body, and can hardly find a path for cluttered environments.

Moreover, all these optimization-based approaches suffer from high computational costs.
Hence, they cannot be applied to efficiently deal with a dynamically changing environment in real time.
Dynamical system (DS) based approaches have been successfully applied to tackle collision avoidance problems in Cartesian space for a point robot or the end-effector of a robotic arm \cite{Mohammad2012DS_modulation}.
A closed-form representation of the velocity field converged at the Cartesian target can be obtained by training a potential field by harmonic functions.
Given a geometric representation of the obstacle in space, the trained potential field can online adapt to the obstacle in space and guide the robot end-effector moving toward the target by following the collision-free velocity vector field. Such approaches are however often restricted to low-dimensional space.

Sampling-based methods can help to alleviate the local minimum issues in these approaches. Such approaches are suitable for searching in high-dimensional spaces and can also be applied to online collision avoidance problems.
One representative approach is the probabilistic roadmap (PRM) \cite{kavraki1996probabilistic}.
It constructs a roadmap in the C-space.
The nodes of the map indicate states, connected by edges.
A path that guides the robot from the initial to the goal state can be obtained by searching the map.

Another category of widely used approaches is proposed based on rapidly exploring random tree (RRT) \cite{lavalle1998rapidly}, which incrementally constructs a space-filling tree by randomly sampling toward unexplored space regions.
Karaman \etal\cite{karaman2011sampling} improved both RRT and PRM by proposing RRT* and PRM*, respectively.
RRT* records each node's distance to its parent and rewires the tree to increase smoothness. 
PRM* uses an adaptive radius to find neighbors and allows all sampled nodes to be connected within a certain distance.
However, both algorithms suffer from dynamic environments.
The roadmaps must be reconstructed once the obstacle moves.

To speed up the convergence rate of sampling-based methods, some combination methods are proposed to incorporate the potential field in sampling-based methods.
However, these approaches only work in Cartesian space since the gradients in joint space are not available.

\subsection{Sliding motion for in-hand manipulation}
Sliding motion is crucial for in-hand manipulation, enabling precise object repositioning and enhancing robotic hand dexterity.
Cherif \etal\cite{cherif2001global} developed a comprehensive approach for the dexterous reorientation of rigid objects, emphasizing the integration of finger-tracking techniques that leverage both rolling and sliding dynamics.
\cite{shi2017dynamic} presents a framework for planning the motion of fingers of a robot hand to create an inertial load on a grasped object for achieving desired in-grasp sliding motion.
These works explore different strategies for in-hand sliding manipulation, primarily focusing on kinematic and control algorithms. Yet, they exhibit limitations in ensuring collision-free motions, particularly in real-time and under uncertain conditions. These studies largely rely on predefined models or heuristics, posing adaptability challenges in dynamic or cluttered environments.

In this paper, we consider a typical in-hand manipulation scenario with a multi-fingered robotic hand grasping an object and trying to adapt to another grasp configuration.
We aim to obtain a collision-free manipulation path or a sliding path for the hand to move from the initial state to the final state, assuming stable object grasp in both states.

We tackle this problem in three steps.
First, in Sec.~\ref{sec::approach}, we learn a model of the collision-free space by estimating a set of distance functions to separate the space region that is collision-free, from the region where collision exists among fingers, the palm, and the object.
Next (Sec.\ref{sec::motion_planning}), we sample within this model for a collision-free path, integrating a reactive control system based on dynamical systems to mitigate unforeseen collisions. 
Finally (Sec.~\ref{sec::exp}), we demonstrate the efficiency of our approach, which combines the reactive control system with sampling-based planning.

\section{NN-based C-space Representation}
\label{sec::approach}
\subsection{Modelling of hand and object}\label{subsec:models}

\subsubsection{Hand model}
We use a 16-DoFs Allegro hand model (Wonik Robotics Co., Ltd., see Fig.~\ref{fig:regrasp_example}) to explain our approach in this work.
The entire hand model (model mesh representation as URDF) consists of 22 geometric bodies, 5 for each finger and 2 for the palm.
For simplification, we categorize these geometric bodies into 5 groups (4 fingers and 1 palm); and we consider the distances between two groups rather than the distances between geometric bodies in the following.
The hand reference frame is positioned at the center of the wrist, with the X-axis points toward the inner of the palm and the Y-axis points toward the ring finger. The Z-axis points from the palm towards the fingertips.

\subsubsection{Object model}
Given the geometric models of the object, we represent the object by a set of points in space, denoted as $\bm X^{o} = \{\bm x_{i}^{o}\}_{i=1}^{N} \in \mathbb{R}^{N \times 3}$, represented in the hand frame. In practice, the point cloud of the object can be obtained using tactile or visual information.
\subsection{C-space representation}
\label{sec::space}
We describe the configuration of the $n$-DoFs robotic hand pose by its joint position vector $\bm q \in \mathbb{R}^{n}$. The corresponding Cartesian position of fingertips can be obtained by forward kinematics $f(\bm{q})$.
If the robotic hand is composed of $m$ geometric bodies, the set of distances between any two detached geometric bodies is given by:
\begin{equation}
    \bm{d}_1(\bm q) = \left\{ h \left(f_{i}(\bm q), f_{j}(\bm{q}) \right)~\Big\vert~i,j\in\{1,2,\dots,m\}, i \neq j\right\},
    \label{eq:c_space:self}
\end{equation}
where $h(\cdot, \cdot)$ outputs the shortest Euclidean distance between the $i$-th and $j$-th geometric bodies. $f_i(\bm{q})$ returns the {mesh positions of the $i$-th geometric body.
We consider the negative distance as penetration depth, a measure of collision.

The set of distances from a point obstacle in Cartesian space $\bm x^o_i \in \mathbb{R}^3$ to the geometric bodies of the robotic hand is calculated as follows:
\begin{equation}
    \bm{d}_2(\bm q, \bm x^{o}_i) = \left\{h \left(f_{i}(\bm q), \bm x^o_i \right)~\Big\vert~i\in\{1,2,\dots,m\}\right\}.
    \label{eq:c_space:obj}
\end{equation}
Then, the minimum distance for a potential collision (either self-collision or hand-object collision) can be found through:
\begin{equation}
    d^{\text{min}}(\bm q, \bm X^o) = \min \left(\bm{d}_1(\bm q), \Bigl\{ \bm{d}_2(\bm q, \bm x_{i}^{o}) \Bigr\}_{i=1,2,\dots,N} \right),
\label{eq:c_space:min_dis}
\end{equation}
where $\left\{ \bm{d}_2(\bm q, \bm x_{i}^{o}) \right\}_{i=1,2,\dots,N}$ is the set of distances from all points of the object to the robot links.

Finally, the C-space $\mathcal{C} \in \mathbb{R}^n $ could be described as:
\begin{subequations} 
\begin{align}
    &\mathcal{C}^o = \{\bm q \in \mathbb{R}^n : d^{\text{min}} < 0\}\\
    &\mathcal{C}^b = \{\bm q \in \mathbb{R}^n : d^{\text{min}} = 0\}\\
    &\mathcal{C}^f = \{\bm q \in \mathbb{R}^n : d^{\text{min}} > 0\}, 
\end{align}\label{eq:c_space}
\end{subequations}
where $\mathcal{C}^o$ denotes the collision region, $\mathcal{C}^b$ represents the collision boundary, and $\mathcal{C}^f$ signifies the collision-free region.

We distinguish between two types of collision: \emph{self-collision} (between fingers or between a finger and the palm), and \emph{hand-object collision} (between a finger or the palm, and the manipulated object). 
We train an neural network (NN) model to detect each type of collision (see Sec.~\ref{sec:NN_1} and Sec.~\ref{sec:NN_2}).
 
\subsection{Self-collision model}
\label{sec:NN_1}
We first explain the training of NN for approximation of $\bm{d}_1(\bm q)$ in Eq.~\eqref{eq:c_space:self}, which calculates the distance between two geometric bodies.
Apart from estimating the distance, it is also necessary to determine whether the robot is in a collision, given its joint configuration.
For this purpose, we train the NN model to output both (1) an estimation of the distance to the obstacle,
and (2) a decision boundary for asserting whether the robot is in a collision.

\subsubsection{Dataset}
To generate the training dataset, we randomly and uniformly sample the hand's joint poses within the joint limits for every joint. Using forward kinematics, we compute the minimum Euclidean distance between two geometrical groups, according to Eq.~\eqref{eq:c_space:self}.
The sampled training dataset is pre-processed to ensure balance for pre-processing of training dataset).

\subsubsection{NN training}
\label{sec:NN1}
We train a four-layer feed-forward NN to estimate the mapping $\bm{q} \rightarrow \bm{d}_1$ with ReLU activation function and a standard mean-square weighted loss:
\begin{equation}
    c_1 = \frac{1}{n_1}\sum_{i=1}^{n_1} w_j (\bm d_{1, i}^{\text{nor}} - \bm d_{1, i}^{\text{pred}})^2,
    \label{eq:NN_1:cost}
\end{equation}
The weights $\bm w = \{w_j\}_{j=1}^3$ are set to different values based on $\bm d_{1, i}^{\text{nor}}$ and $\bm d_{1, i}^{\text{pred}}$ as in Table.~\ref{tab:NN_1:weights}, where $\bm d_{1, i}^{\text{nor}}$ are the normalized distances in $[-1,1]$, and $\bm d_{1, i}^{\text{pred}}$ is the output of the NN. For example, $\bm w = [0.1, 6, 4]$. In this way, the false positives (FP) and false negatives (FN) are penalized, which could improve the accuracy in the binary classification for collision detection and the regression accuracy for distances close to zero.
Since we prioritize safety, accuracy is not a major concern when the prediction distance and ground truth both exceed 0.5, as this indicates a safe distance from collisions. In these instances, we can set $w$ to 0.1 in the cost function to minimize the impact of prediction errors.
A grid search is conducted to determine the optimal structure of the NN model.

\begin{table}[t]
    \caption{Weights for the cost function of self-collision model.}
    \label{tab:NN_1:weights}
    \centering
    \begin{tabular}{cccc}
       \toprule
         & $\bm d_{1, i}^{\text{pred}} \leq 0$ & $0 < d_{1, i}^{\text{pred}} \leq 0.5$   &$ d_{1, i}^{\text{pred}} > 0.5$  \\
            \midrule
          $\bm d_{1, i}^{\text{nor}} \leq 0$ & $w_1$ & $w_2$  & $w_2$ \\
          $ 0< \bm d_{1, i}^{\text{nor}} \leq 0.5$ & $w_3$ & 1 &1 \\
        $\bm d_{1, i}^{\text{nor}} > 0.5$ & $w_3$ & 1 & $w_1$ \\
        \bottomrule
    \end{tabular}
\end{table}

\subsection{Hand-object collision model}
\label{sec:NN_2}
To estimate collisions between the hand and an external object, we train a model of the distance function Eq.~\eqref{eq:c_space:obj}. 
We hence train a total of 3 NNs to determine the distance of the point object to the palm, thumb, and index finger. Notice that the index, middle, and ring fingers use the same NN as these fingers share identical geometric and kinematic models. 

\subsubsection{Dataset}
We generate a uniform distribution of arbitrary points $\bm x^o$ from scaled cuboid hulls (1.1, 2, and 5 times) of hand geometries. We randomly select joint positions, update the pose of each geometric body using forward kinematics, and randomly select $\bm x^o$ from within the resulting cuboid hulls. The distance between each geometric body and a point in space is calculated using the Trimesh library\footnote{\url{https://trimsh.org/}}. 

Three datasets are generated for the palm, the index finger, and the thumb finger, which are denoted as $\bm D_k = \{(\bm q_i, \bm x^o_i), \bm d_{2,i}^r\}_{i=1}^{n_k}, k=2,3,4$, respectively.
The datasets are processed similarly to Sec.~\ref{sec:NN1} for training the NN model.
\subsubsection{NN training}
Three fully-connected NNs are trained to estimate the mappings $(\bm q, \bm x^o) \rightarrow \bm d_2$ for distance estimation. We follow the same training procedure as described in Section \ref{sec:NN1}.

Finally, a NN-based estimation of Eq.~\eqref{eq:c_space:min_dis} is achieved as $\hat{d}^{\text{min}}(\bm q, \bm X^o)$, by combining the self-collision model and the hand-object collision model. 
As the fully-connected NNs of 4 layers are adopted with ReLU, the gradient of the distance with respect to a joint could be derived in realtime.

\section{Motion Planning in Collision-free C-space}\label{sec::motion_planning}
Fig.~\ref{fig:framework} illustrates the framework of our proposed methods. The spatial pose of the obstacle (or object) can be captured in real-time using either a depth camera (represented as point cloud) or a motion capture system (in conjunction with a known geometric model of the object). The trained NNs for both self-collision and hand-object collision models provide real-time estimations of collision distances and gradients. These estimations serve as inputs to the search algorithms (DS-guided RRT* or Dynamic PRM*), which generate collision-free motion paths in the robot's joint space. The robot's joint configuration and the object's pose are continuously used to update the motion paths online.

\begin{figure*}[t]
    \centering
    \includegraphics[width=\textwidth]{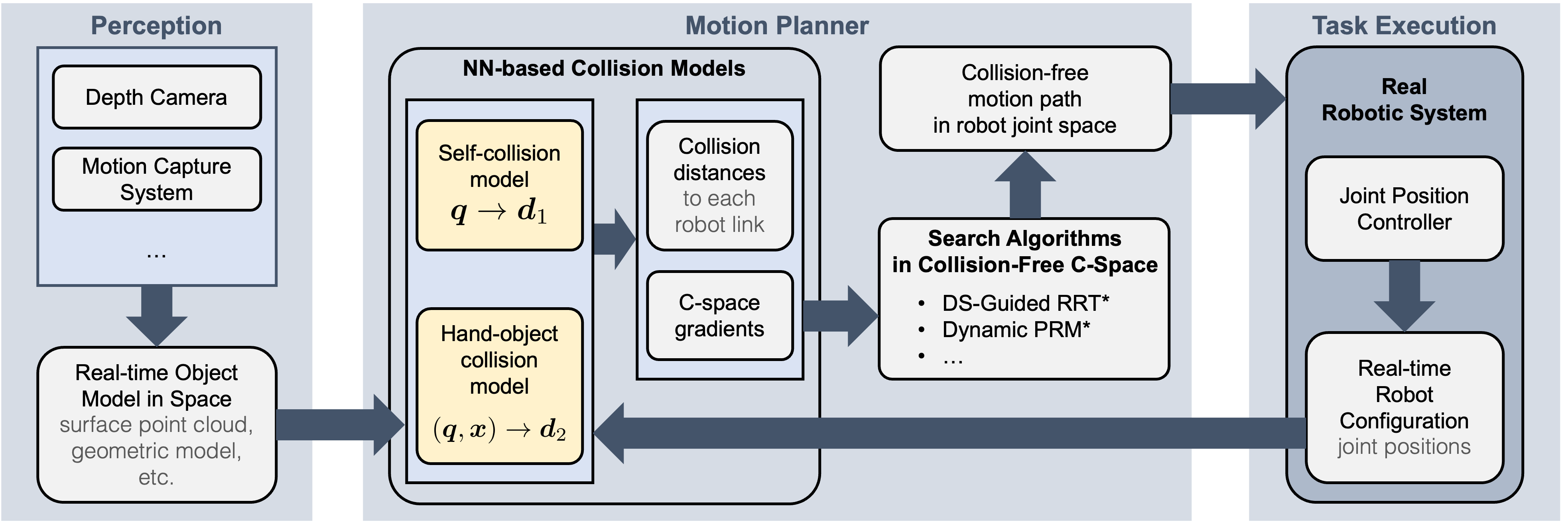}
    \caption{Framework for applying our proposed real-time motion planning algorithms.}
    \label{fig:framework}
\end{figure*}

\subsection{DS-based obstacle avoidance}
\label{sec::DS}
\begin{figure}[t]
     \centering
     \includegraphics[width=0.48\textwidth]{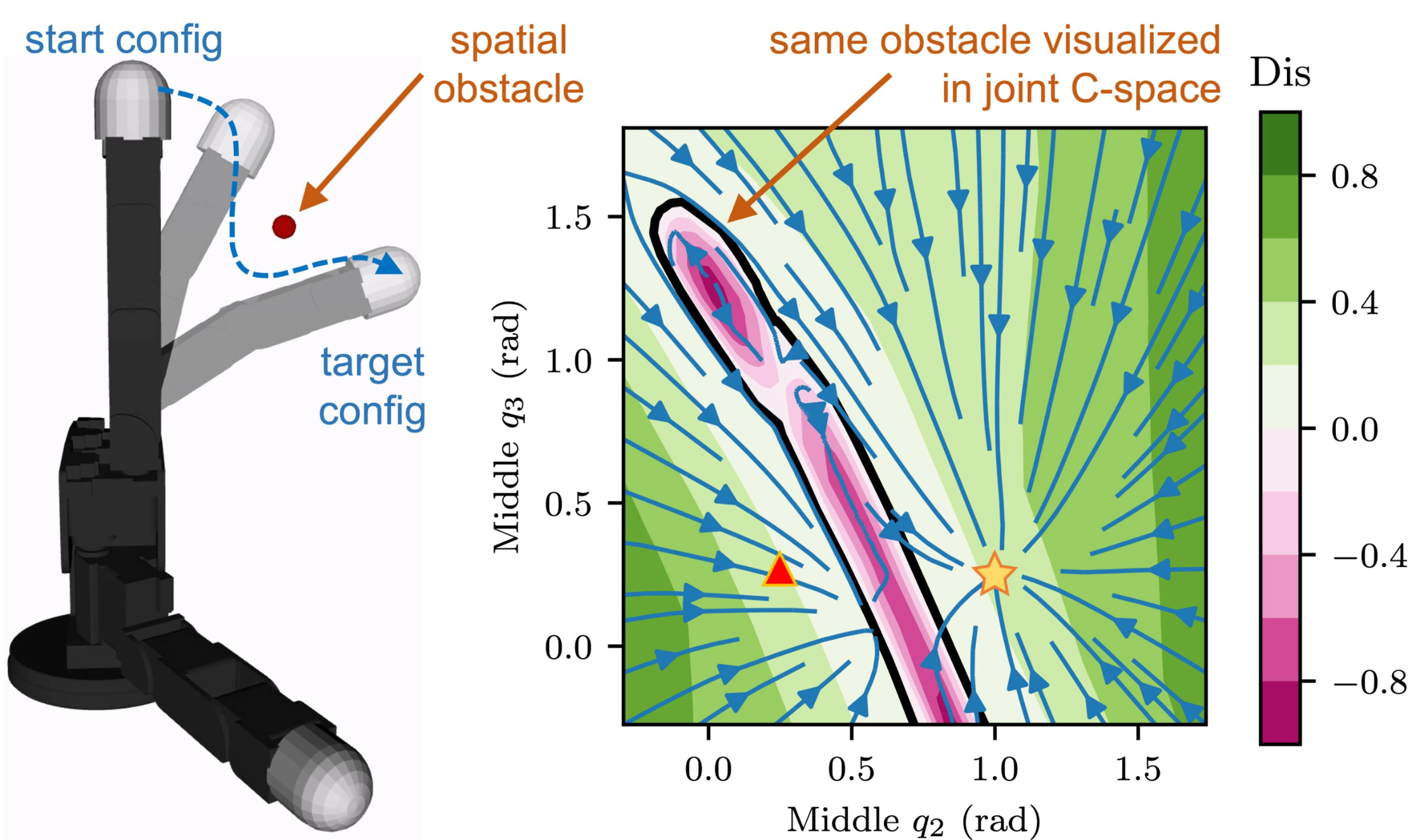}
   \caption{Example of DS-based motion planning.
   The left figure shows the side view of the start (solid, finger straight) and the target (transparent, finger most bend) hand configurations. The middle finger must bend (dashed arrow line) to avoid a spherical obstacle indicated by a red dot.
   The right figure visualizes the isolines and DS streamlines in joint C-space in a 2D space ($q_2$ and $q_3$ of the finger). The black curve is the obstacle boundary. The red triangle and yellow star indicate the start and end configurations in the joint C-space. The color bar refers to the normalized spatial distance between the obstacle and the finger at the corresponding configuration. DS streamlines illustrate the velocity field.
}
\label{fig:DS}
\end{figure}
Inspired by \cite{Mohammad2012DS_modulation}, we build a close-loop control system composed of a linear vector field to the desired joint configuration $\bm q_{g}$, given by a first-order differential equation as $\dot{\bm q} = \bm f(\bm q) = \bm A (\bm q - \bm q_{g})$. 
$\bm A$ is a negative-definite matrix. The dynamics has a single equilibrium point.
Given any current joint $q$, we can calculate the desired velocity command, which will guide the robot converging to $\bm q_{g}$.

The vector field can be deflected by the gradient of the distance to obstacles to avoid them, with larger deflection when closer to the boundary with the collided space.
We first construct a scaled distance function to the collided space boundary using our learned distance function as $\Gamma (\bm q, \bm x) = \alpha_1(d^{\text{min}} - d_{\text{safe}}) + 1$. 
Thus,  $\Gamma>1$ means collision-free.

We compute the analytic gradient of $\Gamma$ with respect to $\bm q$, as $\frac{\partial \Gamma}{\bm q} = \alpha_1 \frac{\partial d}{\partial \bm q}$, by first-order derivation of the NNs decision function. $d_{\text{safe}}$ is a safety margin of the distance. $\alpha_1$ is a positive scaling factor. 
To deflect the dynamics in the vicinity of the collided space boundary, we use the direction formed by the gradient of our distance function. We project onto a basis composed of the tangents and normal to the boundary of the collided space, which is given into matrix $E$, see \cite{Mohammad2012DS_modulation}:
\begin{equation}
     \begin{aligned}          
          \dot{\bm q} &= \bm E \bm D \bm E^{-1} f(\bm q),\\
          \text{with \quad} 
          \bm E &= \left [\frac{\partial \Gamma}{\partial \bm q}, \;\; \bm e^2, \cdots, \; \;\bm e^n \right ], \\
          \bm D &= \text{diag}(\lambda^1, \lambda^2,...,  \lambda^n).   
     \end{aligned}
     \label{DS::modulation}
\end{equation}
The columns of $\bm E$ are composed of $n$ sets of vectors forming a basis of the joint space, with the first column aligned with the normal to the boundary. $\bm D$ is a diagonal matrix, and its eigenvalues are designed as follows:
\begin{equation}
     \left\{ 
\begin{aligned}         
          \lambda^1 &= 1 - \frac{1}{|\Gamma| ^ {\frac{1}{\rho}}} \\
          \lambda^i &= 1 + \frac{1}{|\Gamma| ^ {\frac{1}{\rho}}}\quad 2 \leq i \leq n 
\end{aligned}
\right. ,
\end{equation}
where $\rho >0$ is a reactivity parameter that can be tuned to modulate the sharpness of the deflection around the boundary. 
A decrease in distance $\Gamma$ corresponds to a decrease in $\lambda^1$ and an increase in $\lambda^i$ for $i \geq 2$. Thus, the matrix $\bm E \bm D \bm E^{-1}$ can reduce the velocity along the gradient direction $\frac{\partial \Gamma}{\partial \bm q}$ and increase the velocity along the tangent directions when close to the obstacle. So, the velocity flow vanishes along the normal direction once it is close to the boundary, hence preventing the robot from penetrating the obstacle.

The above approach is guaranteed to have no stable point, except at the desired configuration where the obstacle is convex in C space \cite{Mohammad2012DS_modulation}. When computing deflection in a high-dimensional C-space, it arises often that the boundary displays strong concavities. Hence, it occurs often that the dynamics stops. This is illustrated in Fig.~\ref{fig:DS}. Even if the obstacle is convex (here spherical) in Cartesian space, the obstacle creates a concave shape in C-space. 
Motions that start from the left-bottom region of the obstacle (Fig.~\ref{fig:DS}, right) will become trapped in a local minimum instead of converging towards the attractor.

\begin{algorithm}[t]
     \DontPrintSemicolon
     \caption{\textit{DS-guided RRT*}
     }
     \label{algorithm1}
     \LinesNumbered
     \KwIn{ $\bm q_{a}$, $\bm X^o$, $\bm q_{g}$,}
     \SetKwInOut{Parameter}{Parameters}
     \Parameter{$0 < \alpha_2 < 1$,  $i_\text{max} > 0$}
     \textbf{Initialize}  the tree $\mathcal{T}$ by $\bm q_{a}$,   \newline
     \While {not reaching the goal and $i < i_\text{max}$} {
          update $\bm q_{a}$, $\bm X_{obs}$, $\bm q_{g}$ \label{algorithm1:update}\;
          $\bm d_n( \mathcal{T}_n) := \hat{d}^\text{min}( \mathcal{T}_n, \bm X^o)$  \label{algorithm1:eva_NN}\;
          $\mathcal{T}_n ^f := \{\bm q \in \mathcal{T}_n : \bm d_n(\bm q) > d_\text{safe}   \}$ \;
          $\mathcal{T}_n ^o := \{\bm q \in \mathcal{T}_n :\bm d_n(\bm q) \leq d_\text{safe}   \}$  \label{algorithm1:T^o}\;
          
          $\bm d_e( \mathcal{T}_e) := \hat{d}^\text{min}( \mathcal{T}_e, \bm X^o)$  \label{algorithm1:eva_edge}\;
          remove edges in $\mathcal{T}_e$ where $\bm d_e( \mathcal{T}_e)\leq d_\text{safe} \label{algorithm1:edge_update}$ \;
          a rewire list $\bm Q_\text{rewire} $ := $\{\bm q \in \mathcal{T}_n:$ the node $\bm q$ has different collision state with the last iteration\} \;

          \uIf{no feasible path}{
               $\bm Q_\text{rand}\leftarrow Sample(\mathcal{C})$ \label{algorithm1:sample}\;
               $\bm Q_\text{nearest} \leftarrow Nearest(\bm Q_\text{rand}, \mathcal{T}_n^f)$ \;
               update $\delta  q$ by Eq.~\eqref{rrt::step} for each $\bm q_\text{nearest}$ \;
               $\bm Q_\text{new} := \bm Q_\text{nearest} + \delta \bm q \bm n_q$ \label{algorithm1:Q_new}\;
               get nearby nodes and calculate costs from $\bm q_0$ to $\bm q_\text{new}\in \bm Q_\text{new}$ by $\bm q_\text{near}$\;
               connect all possible $\bm q_\text{new}$ with the best collision-free parents\;
          }
          rewire the nodes nearby $\bm Q_\text{rewire}$ \label{algorithm1:rewire}\;
          Run Algorithm~\ref{algorithm2} with 10\% probability\;
          Robot moves by Algorithm~\ref{algorithm3} \;
          
          $i := i +1$
       }
     \end{algorithm}   
\subsection{DS-guided RRT*}
\label{sec:RRT}
In Fig.~\ref{fig:framework}, we illustrate the framework of our sampling-based methods.The obstacle pointcloud is fed into the NNs for the outputs of distances and gradients, which are sent to the two methods for a collision-free path.
Algorithm.~\ref{algorithm1} illustrates the steps to apply our proposed DS-guided RRT*.
The overall process includes four steps: updating collision states, expanding the RRT tree, rewiring nodes and checking a feasible solution.
Given the current robot joints $\bm q_{a}$ and the goal joints $\bm q_{g}$, the algorithm aims to find a collision-free path under the obstacles $\bm X^o$.
$\mathcal{T}$ refers to the tree, where $\mathcal{T}_n$ are nodes in $\mathcal{C}$ and $\mathcal{T}_e$ are edges. At each iteration, the positions of the obstacle, current robot joints $\bm q_{a}$ and the goal $\bm q_{g}$ are updated (Line~\ref{algorithm1:update}). Then the minimum distances of all nodes are estimated by NNs, with an update of the collision-free nodes $\mathcal{T}_n ^f$ and collided nodes $\mathcal{T}_n ^o$ (Line~\ref{algorithm1:eva_NN}-\ref{algorithm1:T^o}). For each edge in $\mathcal{T}_e$, uniform samples are performed and inputted to NNs for collision check of edges (Line~\ref{algorithm1:eva_edge}), where edges under collision are removed from the tree (Line~\ref{algorithm1:edge_update}).
To handle an environment with moving obstacles, a rewire list is established and populated with nodes whose collision states have changed from the previous iteration. These nodes will be utilized for RRT rewiring.
In case of an infeasible path, tree expansion is necessary to search the space. A fixed number of nodes are randomly selected from $\mathcal{C}$ (Line~\ref{algorithm1:sample}) and their nearest nodes are found in $\mathcal{T}n ^f$.
At each node $\bm q = \bm q_\text{nearest}$ in $\bm Q_\text{nearest}$, a first-order approximation of the distance function is used. To improve the efficiency of RRT exploration, the stepsize is adjusted dynamically with the following constraint:
\begin{equation}
     \alpha_2 \bm d(\bm q_\text{nearest}) + \delta q  \left.\bm n_q ^T \frac{\partial \bm d}{\bm q} \right | _{\bm q = \bm q_\text{nearest}} > d_\text{safe} ,
     \label{rrt::step}
\end{equation}
where the vector $\bm n_q$ represents the normalization of $\bm q_\text{rand} - \bm q_\text{nearest}$.
The update step is bounded $\delta q_l \leq \delta q \leq \delta q_u$ to avoid extreme values.
$\alpha_2 \in [0,1]$ is a constant coefficient. We use $\alpha_2 = 0.5$ in this work. When $\bm n_q$ and the gradient $\left. \frac{\partial \bm d}{\bm q} \right | {\bm q = \bm q_{\text{nearest}}}$ form an acute angle, $\delta q$ is set to the maximum value $\delta q_u$ for greedy exploration. Otherwise, $\delta q$ is reduced following Eq.~\eqref{rrt::step} to allow for more reliable exploration.

The variable stepsize for each $\bm q = \bm q_\text{nearest}$ allows us to create a list of pending nodes $\bm Q_\text{new}$ (Line~\ref{algorithm1:Q_new}). Similar to the RRT* algorithm \cite{karaman2011sampling}, we calculate the cost from the start point to each node $\bm q_\text{new} \in \bm Q_\text{new}$ by evaluating its nearby nodes. The nearby node with the lowest cost is selected as the parent of $\bm q_\text{new}$. The tree is then updated with the new nodes and edges, and an RRT*-like rewiring process is applied to the list $\bm Q_\text{rewire}$ for guaranteed optimality (Line~\ref{algorithm1:rewire}).

In Algorithm~\ref{algorithm2}, we adopt a DS-based solution check that differs from traditional straight-line checks. The list of nodes to be checked is initialized by merging the last feasible nodes $\bm Q_\text{DS}$ and new nodes $\bm Q_\text{new}$ (Line~\ref{algorithm2:init}). Over $K$ iterations of integrating differential Eq.~\eqref{DS::modulation} (Line~\ref{algorithm2:integral_DS}), $\bm Q_0$ progresses to $\bm Q_K$ along the modulated DS. If any nodes approach the goal $\bm q_g$, the original nodes are stored in $\bm Q_\text{DS}$.

Finally, the algorithm~\ref{algorithm3} attempts to find a feasible path. If the list $\bm Q_\text{DS}$ is not empty, indicating that the DS has guided some nodes towards the goal, then a feasible path exists. The node $\bm q_k$ with the lowest cost is selected as an intermediate point towards the goal. If $\bm q_k$ is the root of the tree $\mathcal{T}$, the robot will move along the DS (Line~\ref{algorithm3:along_DS}), otherwise, it will follow the tree (Line~\ref{algorithm3:along_tree}). If $\bm Q_\text{DS}= \varnothing$, the algorithm determines that no feasible path exists and the robot remains stationary (Line~\ref{algorithm3:stay}).

\begin{algorithm}[t]
     \DontPrintSemicolon
     \caption{\textit{Solution check by the integral of DS}
     }
     \label{algorithm2}
     \LinesNumbered
     \KwIn{$\bm Q_\text{new}$, $\bm d$, $\frac{\partial \bm d}{\bm q}$, $\bm q_{g}$,}
     \SetKwInOut{Parameter}{Parameters}
     \Parameter{$0<{\delta t}\leq 0.01 $, $\epsilon=1 e^{-4}$, $K = 50$,  }
     $\bm Q_0 := \{\bm q \in \bm Q_\text{DS} \text{ or } \bm q \in \bm Q_\text{new}\}$ \label{algorithm2:init}\newline
     \For{$i\gets0$ \KwTo $K$}{
          update $\dot{\bm Q}_i$ by Eq.~\eqref{DS::modulation} for every $\bm q \in \bm Q_i$\;
          $\bm Q_{i+1} : = \bm Q_i + \dot{\bm Q}_i {\delta t} \label{algorithm2:integral_DS}$ \;
              $ \bm Q_\text{DS} := \{\bm q \in \bm Q_0 :  \Vert \bm q_{i+1} - \bm q_g \Vert < \epsilon \} $ \label{algorithm2:find_solution}\;
              \uIf{$ \bm Q_\textsc{DS} \neq \varnothing$}{ 
              break
              }
    }
    \Return{$\bm Q_\textsc{DS}$} 
\end{algorithm}

\begin{algorithm}[t]
     \DontPrintSemicolon
     \caption{\textit{Find a feasible path}
     }
     \label{algorithm3}
     \LinesNumbered
     \KwIn{$\mathcal{T}$, $\bm Q_\text{DS}$ }
     \uIf{$ \bm Q_\textsc{DS} \neq \varnothing$}{ 
           find $\bm q_k \text{ with minimum cost to the root by } f_c(\bm q) = \text{cost}(\bm q_0, \bm q), \bm q \in  \bm Q_\text{DS}$ \label{algorithm3:cal_cost}\; 
          \uIf{$f_c(\bm q_k) == 0$}{
               move along DS modulation by Eq.~\eqref{DS::modulation}.
               \label{algorithm3:along_DS}
          }
          \Else{ 
          retrieve the path from $\bm q_0$ to $\bm q_k$ by finding the parent inversely \;
          move toward $\bm q_0$ \label{algorithm3:along_tree}\; 
          \uIf{$\bm q_a$ close to $\bm q_0$}{
               $\bm q_0 := \bm q_1$ \;
               rewire to set $\bm q_0$ as the root \label{algorithm3:update_root}
               }
          }
            feasible path exists
          
     }
     \Else{ 
               stay at $\bm q_0$; no feasible path \label{algorithm3:stay}\;
     }
      
\end{algorithm}  

\subsection{Dynamic PRM*}\label{sec::PRM_star}

Taking advantage of the gradients in C-space, we can update a portion of the nodes in the PRM* online, bringing them closer to the collision boundary. This results in the discovery of a shorter and more feasible path.

Considering one node $\bm q$ in PRM*, the minimum distance to the obstacle boundary and its gradient with respect to $\bm q$ are calculated from NNs. Given this, the node is adjusted to the following one:
\begin{equation}
     \bm q \leftarrow \bm q - \beta_1 \frac{\hat{d}^\text{min} (\bm q) - \beta_2 d_{\text{safe}}  - \frac{\partial \hat{d}^\text{min}}{\partial  \bm  x^o} (\delta \bm x^o)} {\frac{\partial \hat{d}^\text{min}}{\partial \bm q}} ,
     \label{eq:PRM_node_update}
 \end{equation}
where let $0 < \beta_1 < 1$ is the adjustment step coefficient factor. If the obstacle is static or the minimum distance is from the self-collision model, then $\frac{\partial \hat{d}^\text{min}}{\partial \bm x^o} = 0$. This allows the node to update and move in the direction opposite to the gradient until $\hat{d}^\text{min} (\bm q) = \beta_2 d_\text{safe}$. The safety factor $\beta_2 >1$ ensures a secure margin.
Notably, the distance queries and gradient computations are efficiently performed in parallel by NNs, enabling real-time node adjustments through parallel matrix computations of Eq.~\eqref{eq:PRM_node_update}. 
As the nodes are changing under the scenario of moving obstacles, the roadmap is also updated online by a GPU-accelerated k-NN.

\subsection{Motion planning for in-hand sliding manipulation}
\label{sec::sliding}
For in-hand sliding manipulation, a sliding motion can be regarded as a trajectory on the collision boundary $\mathcal{C}^b$. Assuming that the initial and goal joint positions are given and denoted as $\bm q_0$ and $\bm q_n$, respectively. The sampling-based approaches could be adopted to generate a collision-free path $\{ \bm q_i\}_{i=0}^n$, which is then projected onto the obstacle boundary $\mathcal{C}^b$ by using Eq.~\eqref{eq:PRM_node_update} with $d_{\text{safe}} = 0$. Several iterations are made to make the path close enough to $\mathcal{C}^b$.

\section{Experimental Evaluation}\label{sec::exp}
We validate our proposed algorithms both in simulation and on a real robotic hand.
The NN-based collision detector was run on an Intel Core i7-12700HQ CPU and RTX3070ti GPU.
The Allegro robotic hand was connected to an Ubuntu 20.04 laptop and controlled by a joint position PID controller in ROS Noetic at 200 Hz.

\subsection{Pre-processing of training dataset for self-collision NN model}\label{appendix:processing}
A critical point when generating such a dataset is balancing the collision and collision-free classes.
We balance the dataset by ensuring that only half of the samples are collision-free.
It takes 3 hours to generate 2 million samples with an Intel Core i7-12700HQ CPU (20 processes).
The dataset is denoted as joint-distance pairs $\bm D_1 = \{ \bm q_i, \bm d_{1,i}^r\}_{i=1}^{n_1}$ with $n_1$ samples and $\bm d_{1,i}^r \in \mathbb{R}^{10}$.

To improve the accuracy and convergence speed of the NN, the dataset is scaled by min-max normalization for positive and negative distances separately, which is written as:
\begin{equation}
     \bm d_{1, i}^{\text{nor}} = \left\{
     \begin{matrix}
          & {\bm d_{1,i}^r} / {\bm d_{1}^{\text{up}}} , \quad \bm d_{1,i}^r \geq 0 \\
          & {\bm d_{1,i}^r } / {\bm d_{1}^{\text{low}}} ,\quad \bm d_{1,i}^r < 0
     \end{matrix}
     \right. .
     \label{eq:NN_1:pre}
\end{equation}
Rather than using the minimum and maximum distances for normalization, we established the upper bound $\bm d_{1}^{\text{up}}$ and lower bound $\bm d_{1}^{\text{low}}$ empirically, based on the distributions of each column in the dataset.

\begin{figure}
\centering
\includegraphics[width=0.48\textwidth]{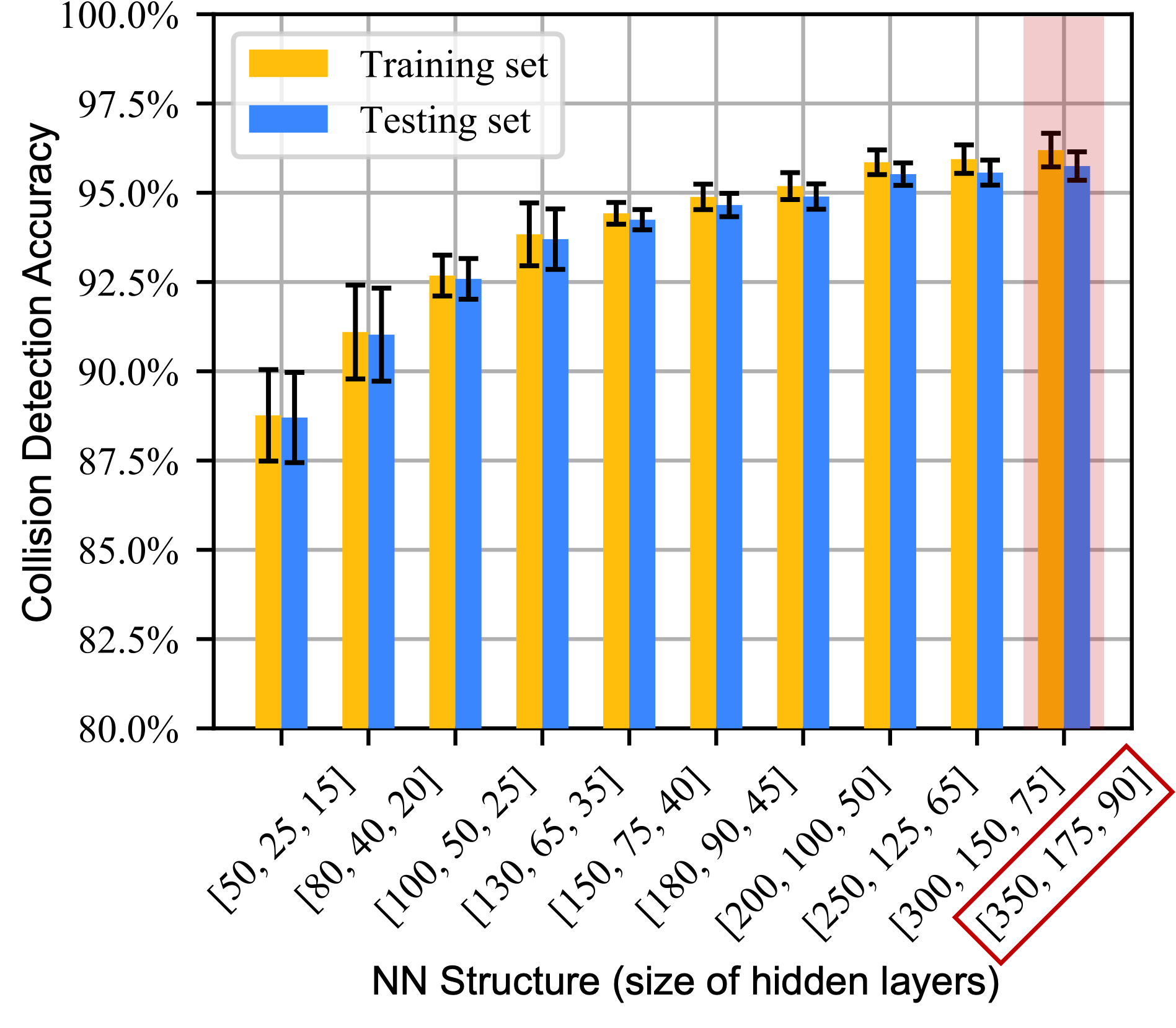}
\caption{
Evaluating the Accuracy of Neural Networks with Different Architectures through Cross-Validation. The bar graph displays the mean accuracy of 20 trials of various neural network structures, each trained for 10,000 epochs. The height of each bar indicates the mean accuracy, while the error bars represent the standard deviation. The X-axis labels denote each neural network structure's size (number of neurons).
The performance no longer significantly improves for NN topologies larger than [350, 175, 90], which is hence selected in designing our neural network.}
\label{fig:NN_1:cv}
\end{figure}

\subsection{Evaluation of self-collision NN model}\label{appendix:evaluation_self}

\begin{figure}
\centering
\includegraphics[width=0.4\textwidth]{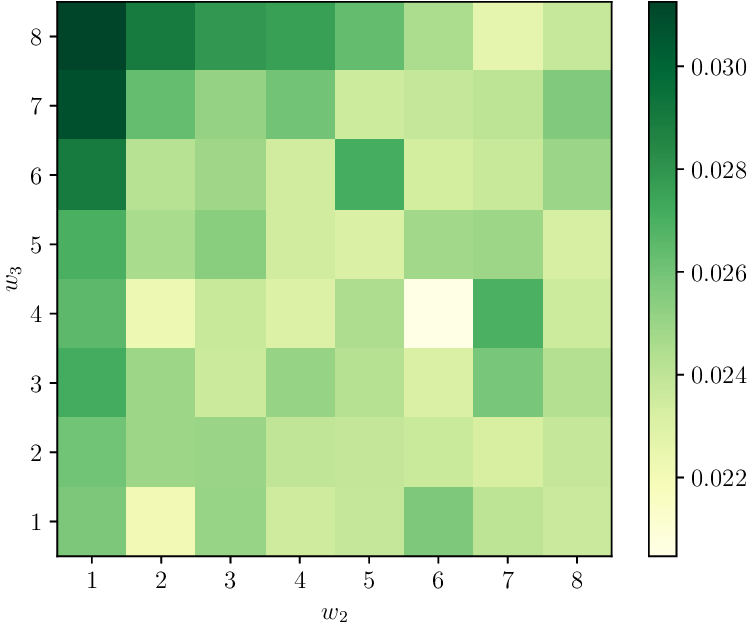}
\caption{
Heatmap for a grid search on the false positive rate (FP). The goal is to determine the optimal values of the weights $w_2$ and $w_3$ in the cost function given by Eq.~\eqref{eq:NN_1:cost}. The $(w_2, w_3)$ pair with the lowest FP is highlighted in white $(w_2=6, w_3=4)$, and used in our self-collision model.
}
\label{fig:NN_1:heatmap}
\end{figure}

To evaluate our model's performance, we divide our dataset into two parts: $50\%$ for training and $50\%$ for testing, each containing 1 million samples. The training process involves 20,000 iterations to optimize the network's parameters.

Fig.~\ref{fig:NN_1:cv} illustrates the cross-validation results with increasing layer widths. NNs with 2 layers and 4 layers are also tested. The accuracies are not significantly improved after the layer widths of $[350,175,90]$. To determine the weights in Eq.~\eqref{eq:NN_1:cost}, we performed a grid search for $(w_1, w_2)$ (see Fig.~\ref{fig:NN_1:heatmap}).

\begin{table}[t]
    \newcommand{\tabincell}[2]{\begin{tabular}{@	{}#1@{}}#2\end{tabular}}
    \caption{
    Evaluation of the self-collision model presented in terms of False Positives (FP) and False Negatives (FN), where the \textit{positive} class is the collision-free state, and the \textit{negative} one indicates the presence of self-collision.
    }
    \label{tab:NN_1:validation}
    \centering
    \begin{tabular}{cccccc}
        \toprule
        \tabincell{c}{Collision\\ detection\\ accuracy }
            & FP & FN & \tabincell{c}{Regression\\ error (mm)} & \tabincell{c}{FP with\\ 3 mm \\margin} & \tabincell{c}{FP with\\ 5 mm\\ margin} \\
            \midrule
        96.5\% & 2.61\% & 4.12\% & 0.63 (0.94) & 0.22\% & 0.029\% \\
        \bottomrule
    \end{tabular}
\end{table}

Then we run the validation on another 1 million samples. To get back the real distances, we invert Eq.~\eqref{eq:NN_1:pre}. Table.~\ref{tab:NN_1:validation} shows the collision detection accuracy.
We evaluate the sensitivity of our distance function in determining the minimum safety distance to the boundary that minimizes false positives (FP), i.e., poses incorrectly classified as non-collided, across collided and free space.
When the safety margin of the distance $d_{\text{safe}}$ is $0$, the ratio of FP is $2.61$\%. For $d_{\text{safe}}$ equal to $3$mm or $5$mm, the FP decreases significantly. We hence set such margin for our experiments (See Fig.~\ref{fig:NN_1:c_space}).

\begin{figure}
    \centering
    \begin{subfigure}[b]{0.5\textwidth}
        \centering
        \begin{subfigure}[b]{0.3\textwidth}
            \includegraphics[scale=1]{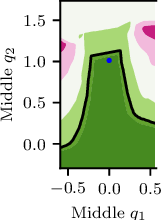}
            \caption*{}
        \end{subfigure}
        \hspace{-0.1cm}
        \begin{subfigure}[b]{0.3\textwidth}
             \includegraphics[scale=1]{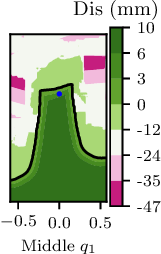}
             \caption*{}
         \end{subfigure}
         \begin{subfigure}[b]{0.3\textwidth}
             \includegraphics[scale=1]{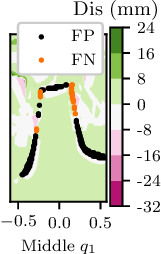}
             \caption*{}
         \end{subfigure}
     \end{subfigure}

     \vspace{-0.3cm}
     \begin{subfigure}[b]{0.5\textwidth}
          \centering
          \begin{subfigure}[b]{0.3\textwidth}
              \includegraphics[scale=1]{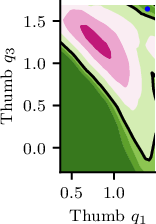}
              \caption{NN prediction}
          \end{subfigure}
          \hspace{-0.1cm}
          \begin{subfigure}[b]{0.3\textwidth}
               \includegraphics[scale=1]{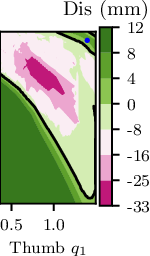}
               \caption{Ground truth}
           \end{subfigure}
           \begin{subfigure}[b]{0.3\textwidth}
               \includegraphics[scale=1]{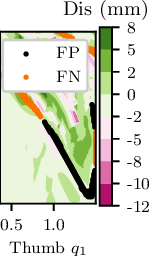}
               \caption{Error}
           \end{subfigure}
       \end{subfigure}
     \caption{C-space visualization under the fully-closed hand configuration with distance isolines.
     \emph{(a) 1st column}: The C space calculated by the NN.
     \emph{(b) 2nd column}: The C space of groundtruth.
     \emph{(c) 3rd column}: Distance prediction error.
     \emph{1st row}: The 1st and 2nd joints of middle finger.
     \emph{2nd row}: The 1st and 3rd joints of Thumb finger.
     The blue point refers to the current robot joints and the black curve is the collision boundary. The predicted C space is almost the same as the ground truth. FP and FN samples occur nearby the boundary. Thus, increasing the safety margin could reduce the FP rate.
     }
     \label{fig:NN_1:c_space}
\end{figure}

\subsection{Evaluation of hand-object collision NN model}\label{appendix:evaluation_hand_object}
\begin{table}[t]
    \newcommand{\tabincell}[2]{\begin{tabular}{@	{}#1@{}}#2\end{tabular}}
    \caption{Evaluation of the hand-object collision model. The collision-free state is referred to as the positive class, while the negative class indicates the existence of a collision.}
    \label{tab:NN_2:validation}
    \centering
    \begin{tabular}{cccccc}
       \toprule
        &\tabincell{c}{Collision\\ detection\\ accuracy }
            & FP &FN & \tabincell{c}{Regression\\ error (mm)} &  \tabincell{c}{FP with\\ 5 mm\\ margin} \\
            \midrule
            Palm & 98.0\% & 0.23\% & 2.3\% & 0.74 (0.63) &  0 \\
            Index & 96.4\% & 2.9\% & 3.7\% & 0.86 (0.75) &  2.8 e{-6} \\
            Thumb & 95.4\% & 2.9\% & 5.3\% & 1.02 (0.97) &  8.9 e{-6} \\
        \bottomrule
    \end{tabular}
\end{table}

The evaluation results are shown in
Table.~\ref{tab:NN_2:validation}. The accuracy of the NN for the index and thumb fingers is comparable to that of the palm model (98\%), which has the highest accuracy due to its static nature that is not impacted by joint movements.
For the three NNs, FP rates reduce to a marginal number when using 5 mm as the safety margin. 
We denote the three NNs distance functions as $\hat{\bm d}_k(\bm q, \bm x^o), k = 2,3,4$ for the distances between the point $\bm x^o$ and the palm, index, and ring finger, respectively.

\subsection{Comparisons of collision detection}\label{sec::comparison}
We use the FCL (Flexible Collision Library) as a baseline for collision detection. The computational performance is investigated and the results are shown in Table.~\ref{tab:collision_comparison}. Considering the Allegro hand with a random joint configuration in 16D and a random point in the workspace, both self-collision and hand-point collision are counted in the time cost.

For batch sizes greater than or equal to 100, FCL is executed by 20 processes.
The FCL time cost increases almost linearly along the batch size, while the NN-based method shows a significantly faster detection. One reason is that FCL needs to compute the forward kinematics for updating poses for all geometric bodies, and then calculate the minimum distance, which is computationally intensive. NN-based collision detection can achieve a query interface of over 100 Hz for 10,000 samples.

\begin{table}[t]
    \newcommand{\tabincell}[2]{\begin{tabular}{@	{}#1@{}}#2\end{tabular}}
    \caption{Comparison of computational time cost (in milliseconds) for different methods of collision detection.}
    \label{tab:collision_comparison}
    \centering
    \begin{tabular}{lccc}
       \toprule
           Sample size & FCL
            & NN by CPU   & NN by GPU  \\
            \midrule
           1 & 11.0 &       \textbf{1.3} &    5.4           \\
           10 & 110.2 &     \textbf{1.7} &  5.4  \\
           100 & 238.4 &   \textbf{2.8} &   5.8  \\
           1,000 & 1167.3 &  8.3  &   \textbf{5.8}\\
           10,000 & 11448.5 & 31.4 &  \textbf{8.5} \\
           100,000 & 122005.7 & 458.5 & \textbf{45.7}   \\
           1,000,000 &  1189485.1       & 4531.8      &  \textbf{412.4} \\
        \bottomrule
    \end{tabular}
\end{table}

\subsection{Simulations of collision-free motion planning}\label{sec::simulation_sampling}

We compare our motion planning approaches with the following commonly used approaches:
\begin{enumerate}[(a)]
    \item RRT*~\cite{karaman2011sampling} combines the original RRT method with rewiring to generate optimal paths.
    \item P-RRT*~\cite{qureshi2016potential} uses a potential field approach to guide sampling in RRT*.
    \item RRT* using a variable stepsize as in Eq.~\eqref{rrt::step}.
    \item PRM*~\cite{karaman2011sampling}.
    \item iLQR ~\cite{todorov2005generalized}.
\end{enumerate}
Metrics include success rate, computation time, and the length of a feasible path. Both \emph{static} and \emph{dynamic} environments are tested in 2D and 16D spaces. For RRT-based methods, maximum nodes $i_\text{max} = 1000$. In PRM-based methods, the number of samples is set as 200. 
All methods are able to use the learned distance functions for collision detection, while our methods exploit the gradients of the distance functions.

First, considering a 2D and static scenario as Fig.~\ref{fig:DS}, a feasible path from DS-guided RRT* is shown in Fig.~\ref{fig:2D:DS}, where the path consists of several RRT-based straight lines (in red) and a DS-guided curve (in blue) to the goal.
At the junction point, if a solution check by a straight line to the goal is tried, the collision checker will return the collision state, and thus more RRT samples are required.
Comparisons with metrics are recorded in Tab.~\ref{tab:2D:validation} with 100 trials for each. All the RRT-based methods have a 100\% success rate. The DS-guided RRT* does not return an optimal path and has a longer path length than others.
However, it shows a faster computation time than other RRT methods thanks to the variable stepsize and the solution check by DS.
Because of the narrow passage, PRM* shows a lower success rate, which is slightly improved by the dynamic PRM*. Taking advantage of parallel computing, PRM-based methods are much faster than RTT-based ones, where the computation time includes updating nodes, building the roadmap, and finding a feasible path by PRM*. 
However, as an example of gradient-based optimization methods, iLQR could not find a feasible a path due to local minimum, as shown in the last rows of Tab.~\ref{tab:2D:validation} and \ref{tab:16D:validation}.

\begin{figure}[t]
\centering
     \includegraphics[width=0.48\textwidth]{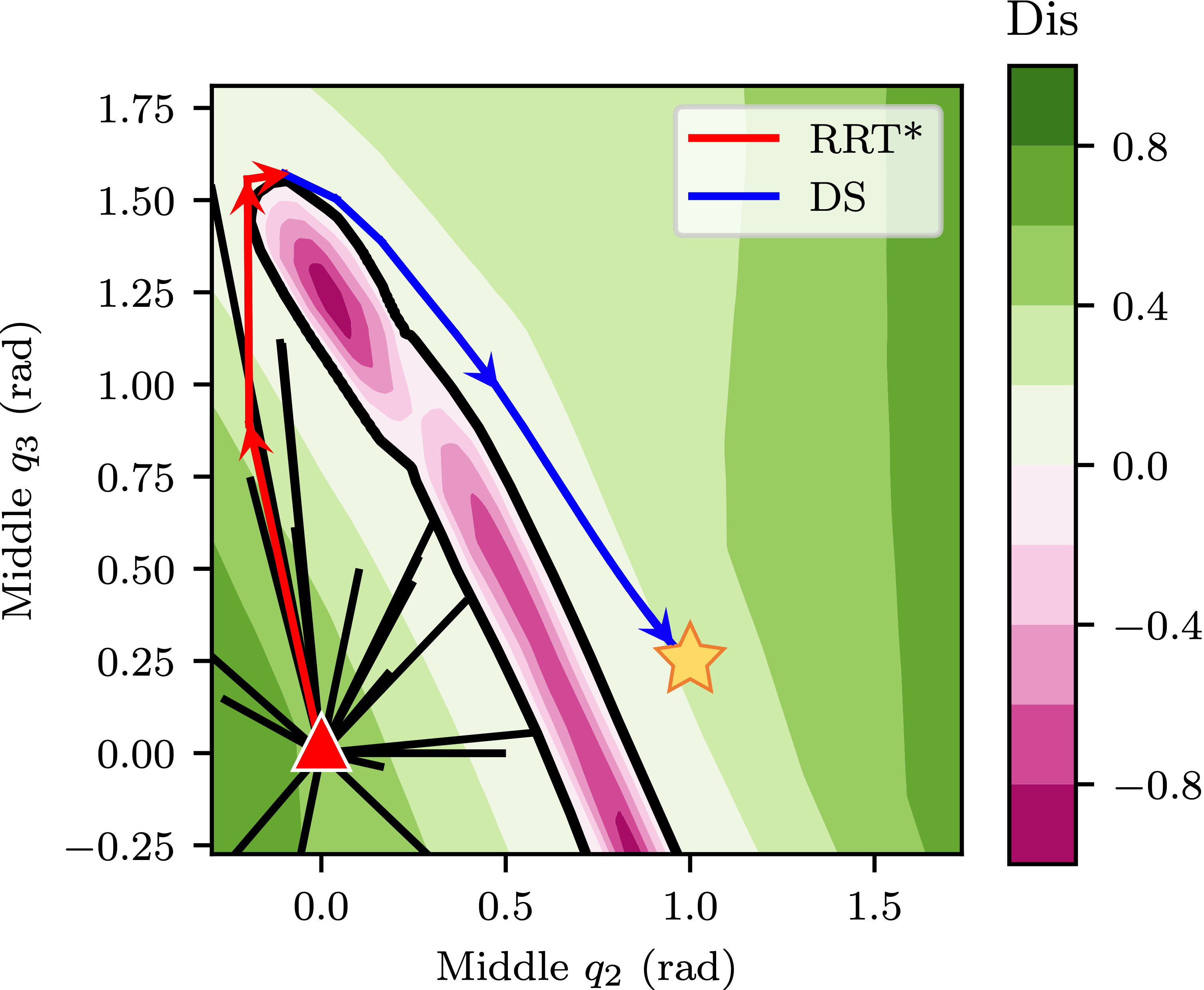}
  \caption{
  An example of a DS-guided RRT* search in C-space projected 2D space of ($q_2$, $q_3$). The black line segments represent the sampled paths, while the feasible path from the start (red triangle) to the goal (yellow star) is a combination of the straight lines generated by RRT* (red arrow path) and the curve from DS (blue arrow path).
  }
    \label{fig:2D:DS}
\end{figure}

\begin{figure}[t]
    \centering
    \includegraphics[width=0.45\textwidth]{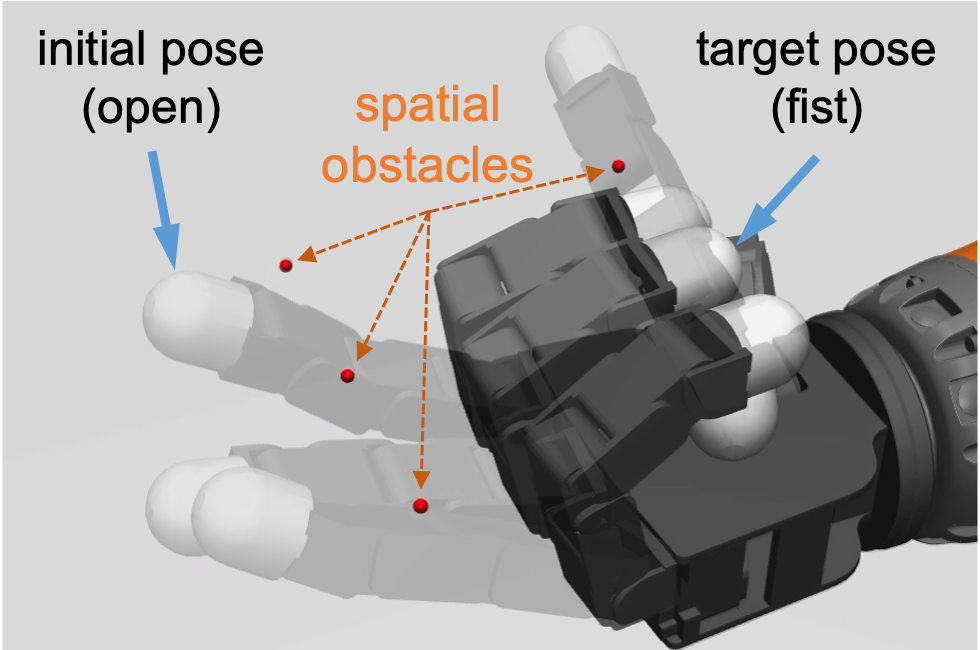}
    \caption{Closing a 16-DoF fully-opened hand while avoiding (1) collisions with four spatial obstacles represented by red dots, and (2) collisions between the fingers.}
    \label{fig:mujoco_16D}
\end{figure}

We then evaluate our approach for full-hand collision avoidance in a 16D space (see Fig.~\ref{fig:mujoco_16D}).
The goal is to clench the hand from \emph{open} pose to the \emph{fist} pose. There are four spherical obstacles on the way of each finger. Both self-collision and hand-object collision should be taken into account, with narrow passages for motions. 
The simulation results is shown in Tab.~\ref{tab:16D:validation}.
Traditional RRT-based methods suffer from high dimension problems and have a very poor success rate, while the DS-guided RRT* is able to keep the 100\% success rate but still returns a longer path. On the other hand, dynamic PRM* offers a shorter path than PRM*. The computation time is not affected by the high dimension. 
Compared to the work by \cite{yeh2012uobprm}, where the gradients are not available and nodes are sampled by intersections between random segments and C-obstacles, the dynamic PRM* takes about 0.15 s for computation, rather than a few seconds. 

For dynamic environments, the DS-guided RRT* and dynamic PRM* are able to update the collision states for all nodes and query a path within 30 ms. Note that both methods are suitable for multi-query tasks, while other methods need to compute a path by full initialization, and the time cost is equal to the computation time in Tab.~\ref{tab:2D:validation} and Tab.~\ref{tab:16D:validation}.

\begin{table}[t]
    \newcommand{\tabincell}[2]{\begin{tabular}{@	{}#1@{}}#2\end{tabular}}
    \caption{Comparison of computational performance of different search algorithms in a 2D static case illustrated by Fig.~\ref{fig:DS}.}
    \label{tab:2D:validation}
    \centering
    \begin{tabular}{cccc}
       \toprule
        & \tabincell{c}{Success \\ rate}
            & \tabincell{c}{Computation\\ time (s)} & Path length (rad)    \\
            \midrule
            RRT* &  1 &  1.56 (1.16) &  3.73 (0.17)\\
            P-RRT*~\cite{qureshi2016potential} & 1 & 1.88 (0.87) & 3.70 (0.14)\\
            \tabincell{c}{RRT* with \\ variable step}  & 1 & 1.06 (0.57)  & 3.82 (0.24)\\
            \textbf{DS-guided RRT*} & 1 & 0.99 (0.52) & 4.06 (0.74)  \\
        PRM*~\cite{karaman2011sampling} & 0.55 & 0.10 (0.03) & 3.66 (0.06) \\
        \textbf{Dynamic PRM*} & 0.7 & 0.17 (0.03) & 3.65 (0.05) \\
        iLQR & 0 & - & -\\
        \bottomrule
    \end{tabular}
\end{table}

\begin{table}[t]
    \newcommand{\tabincell}[2]{\begin{tabular}{@	{}#1@{}}#2\end{tabular}}
    \caption{Comparison of computational performance of different search algorithms in a 16D static case.}
    \label{tab:16D:validation}
    \centering
    \begin{tabular}{cccc}
       \toprule
        & \tabincell{c}{Success \\ rate}
            & \tabincell{c}{Computation\\ time (s)} & Path length (rad)    \\
            \midrule
            RRT*,  P-RRT*   & 0.02  &  - & -  \\
            \tabincell{c}{RRT* with \\ variable step}   &  0.17& 47.81 (12.95)   & 6.33 (0.59)  \\
           \textbf{ DS-guided RRT*} & 1 & 1.61 (1.45)  &  13.28 (3.5)  \\
        PRM*~\cite{karaman2011sampling} & 0.8 &0.15 (0.01) & 8.64 (1.77) \\
        \textbf{Dynamic PRM*} & 0.8 & 0.15 (0.00) & 8.47 (1.56) \\
         iLQR & 0 & - & -\\
        \bottomrule
    \end{tabular}
\end{table}

\subsection{Evaluation on a real robotic hand}\label{sec::real_robot}
We evaluated our proposed approaches in four experimental scenarios with a left Allegro robotic hand.
The robotic hand was stably mounted upright on a table surface through a stand, such that the fingertips point upwards when the hand is fully opened (see Fig.~\ref{fig:regrasp_example}).
$\text{OptiTrack}^{\text{\texttrademark}}$ motion capture system and an Intel RealSense D435 camera were used for spatial localization.
Four markers were attached to the base of the hand to localize the hand reference frame.
The OptiTrack motion capture system guarantees an accuracy of less than $1$ mm, while for the pose estimation of the YCB objects and the rope, the error margin of the pose tracking by an RGBD camera is within $4$ mm. Thus, we set the safety margin  $d_{\text{safe}}$ as $5$ mm.

\subsubsection{Reactive collision avoidance}
The first experimental scenario aims to evaluate the performance of our proposed algorithm for real-time collision avoidance in 16D.
We used a glue stick (cylindrical shaped, radius 8mm, length 147mm) as the experimental object.
Three markers were attached to the object surface to track its pose in real-time.
The glue stick was represented by 100 spheres along its axis with a radius of 8mm. 
Spatial positions were represented in the hand frame. 
Fig.~\ref{fig:reactive_snapshots}(a) illustrates the experimental setup and the initial state of the robotic hand.
The experimental object was attached to the end of a long stick, held by a human experimenter, moved toward the hand and attempted to collide with the index finger (Fig.~\ref{fig:reactive_snapshots}(a)).
The spatial position and orientation of the object were detected by the motion capture system in real-time.
A collision is about to happen and needs to be avoided, if (1) the predicted distance between the experimental object and the hand, or (2) the distance between any pair of two fingers, reduces below the safety margin.
In this case, the hand performed an adjustment motion along the direction of $\frac{\partial \hat{d}^\text{min}}{\partial \bm q}$, which indicates the fastest direction to increase the distance $\hat{d}^\text{min}$ and to avoid a collision (see Fig.~\ref{fig:reactive_snapshots}(b),(c)).

In a second experimental trial (Fig.~\ref{fig:reactive_snapshots}(d),(e),(f)), the experimenter held the stick and tried to collide with all three fingers from the back (Fig.~\ref{fig:reactive_snapshots}(d)), and then adjusted pose to approach the thumb (Fig.~\ref{fig:reactive_snapshots}(e)).
Notice that in Fig.~\ref{fig:reactive_snapshots}(e) and (f), both the middle and the ring fingers moved away from the object, even though the object did not approach them. This is because the index finger moved to avoid a potential collision with the object, causing the distance between adjacent fingers to reduce. Hence, all fingers moved to avoid self-collisions, i.e., collisions between adjacent fingers, guided by the gradient from our self-collision model.

\begin{figure}[t]
    \centering
    \begin{subfigure}[b]{\textwidth}
         \begin{subfigure}[b]{0.15\textwidth}
            \includegraphics[width=\textwidth]{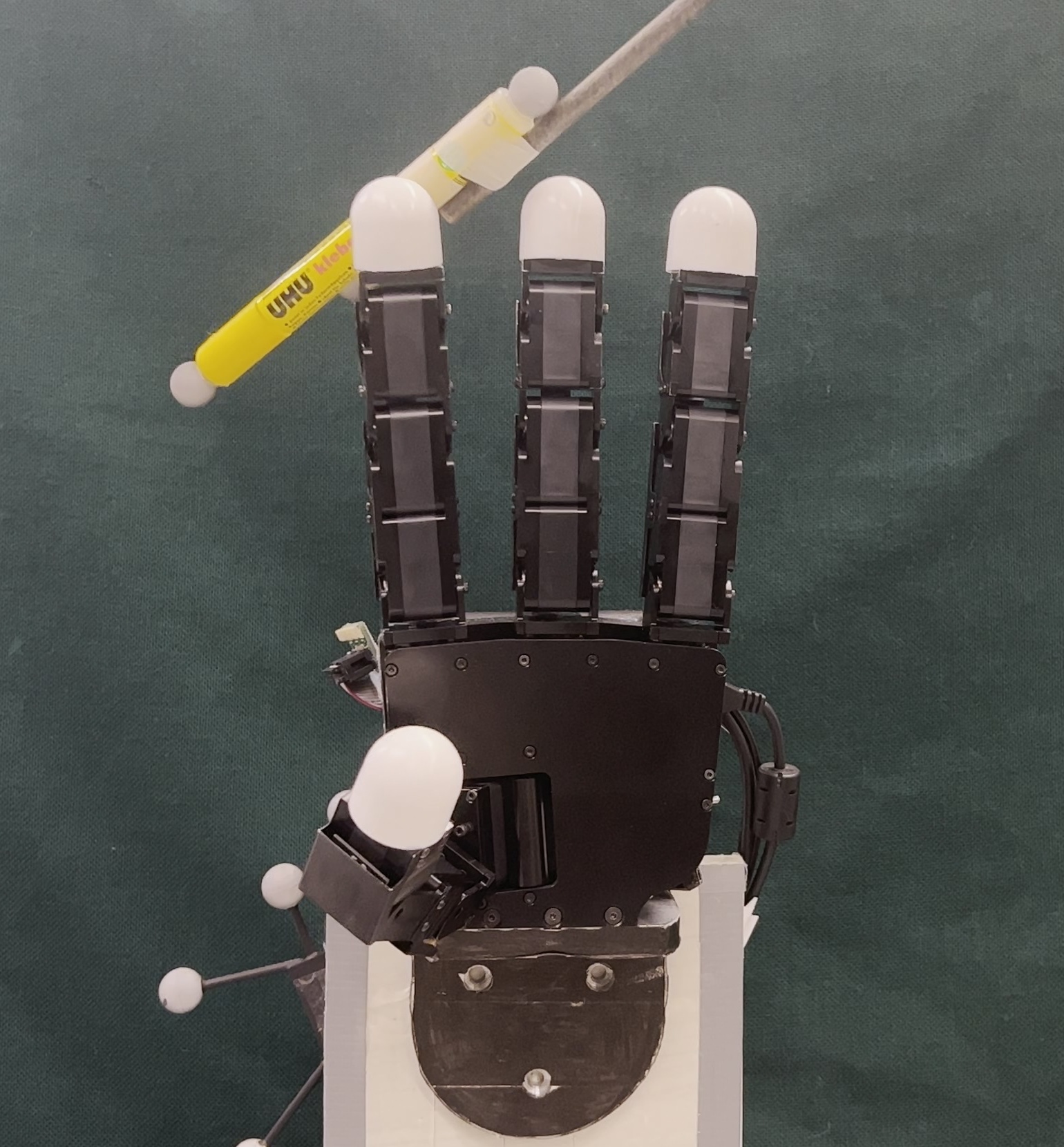}
            \caption{}
        \end{subfigure}
        \begin{subfigure}[b]{0.15\textwidth}
            \includegraphics[width=\textwidth]{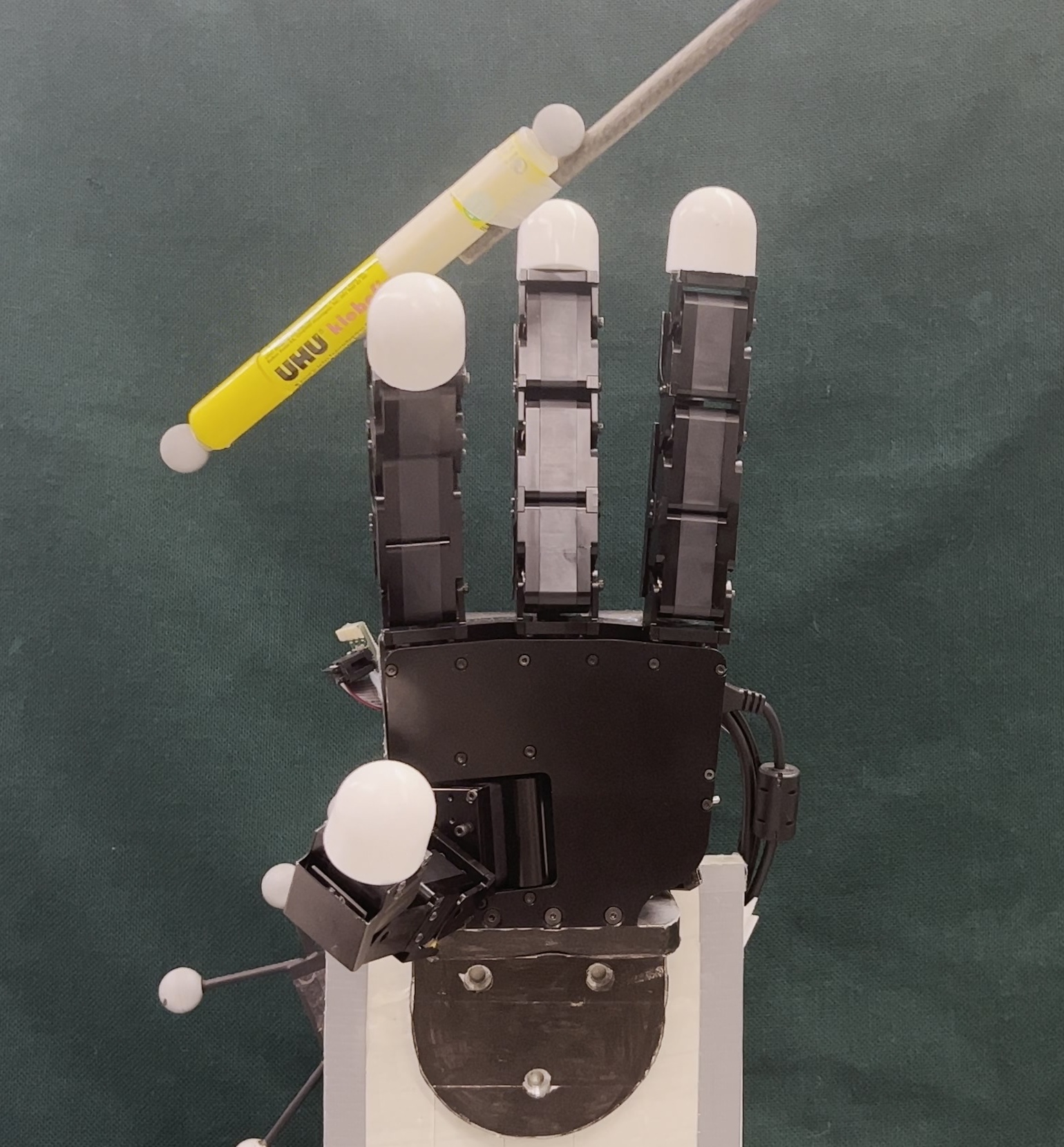}
            \caption{}
        \end{subfigure}  
        \begin{subfigure}[b]{0.15\textwidth}
            \includegraphics[width=\textwidth]{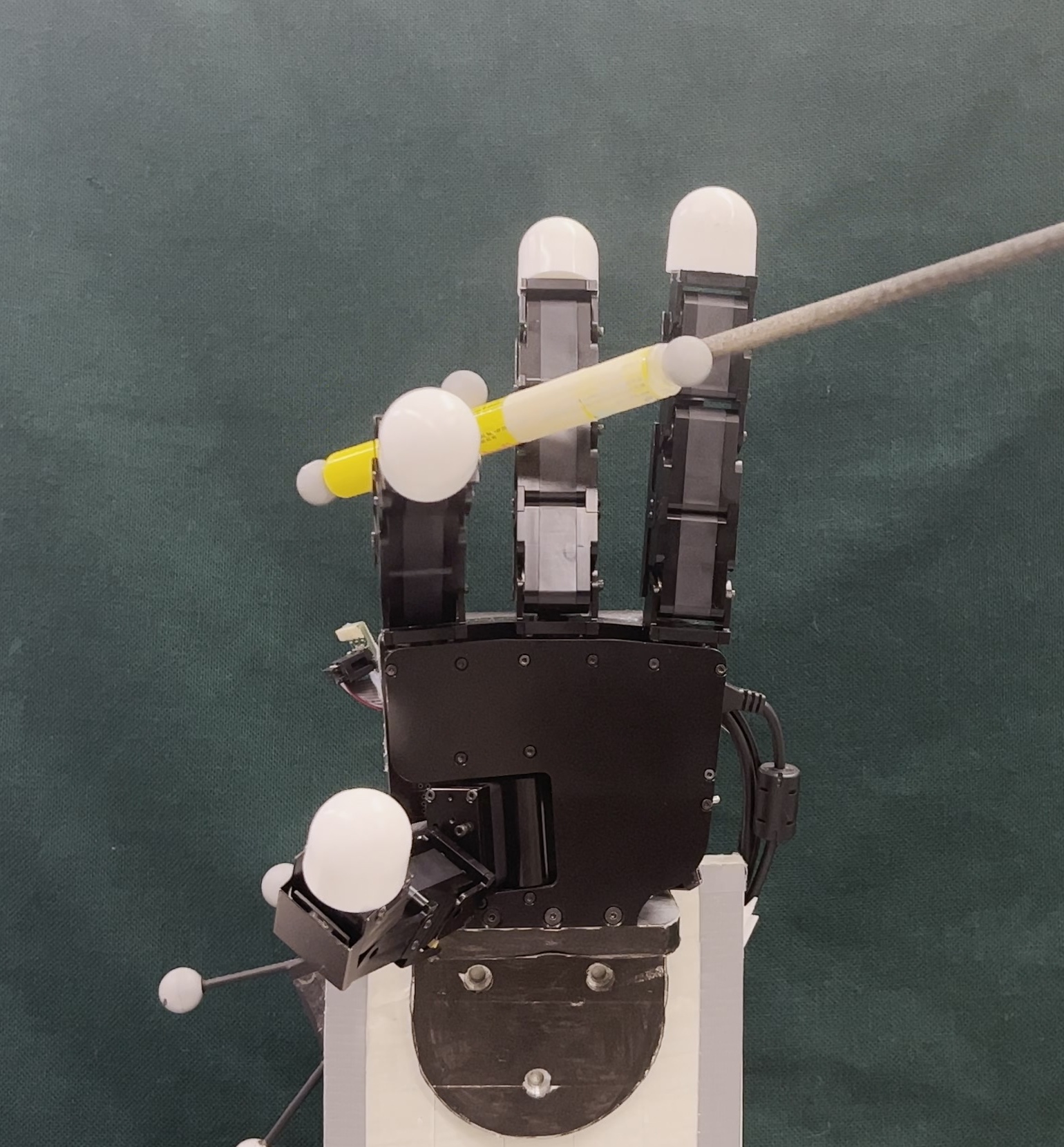}
            \caption{}
        \end{subfigure}  
    \end{subfigure}
    
    \vspace{0.3cm}
    \begin{subfigure}[b]{\textwidth}
         \begin{subfigure}[b]{0.15\textwidth}
            \includegraphics[width=\textwidth]{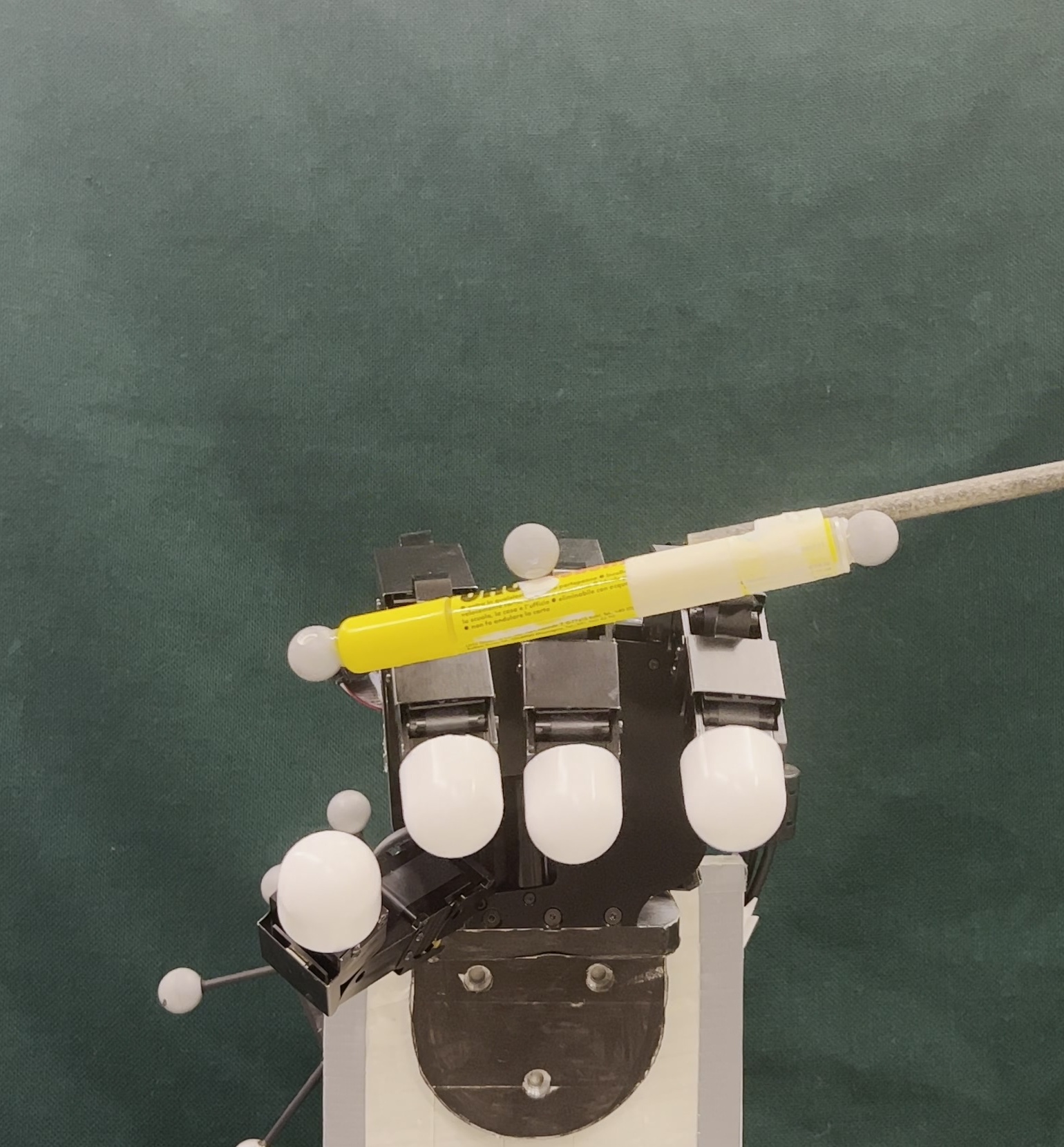}
            \caption{}
        \end{subfigure}
        \begin{subfigure}[b]{0.15\textwidth}
            \includegraphics[width=\textwidth]{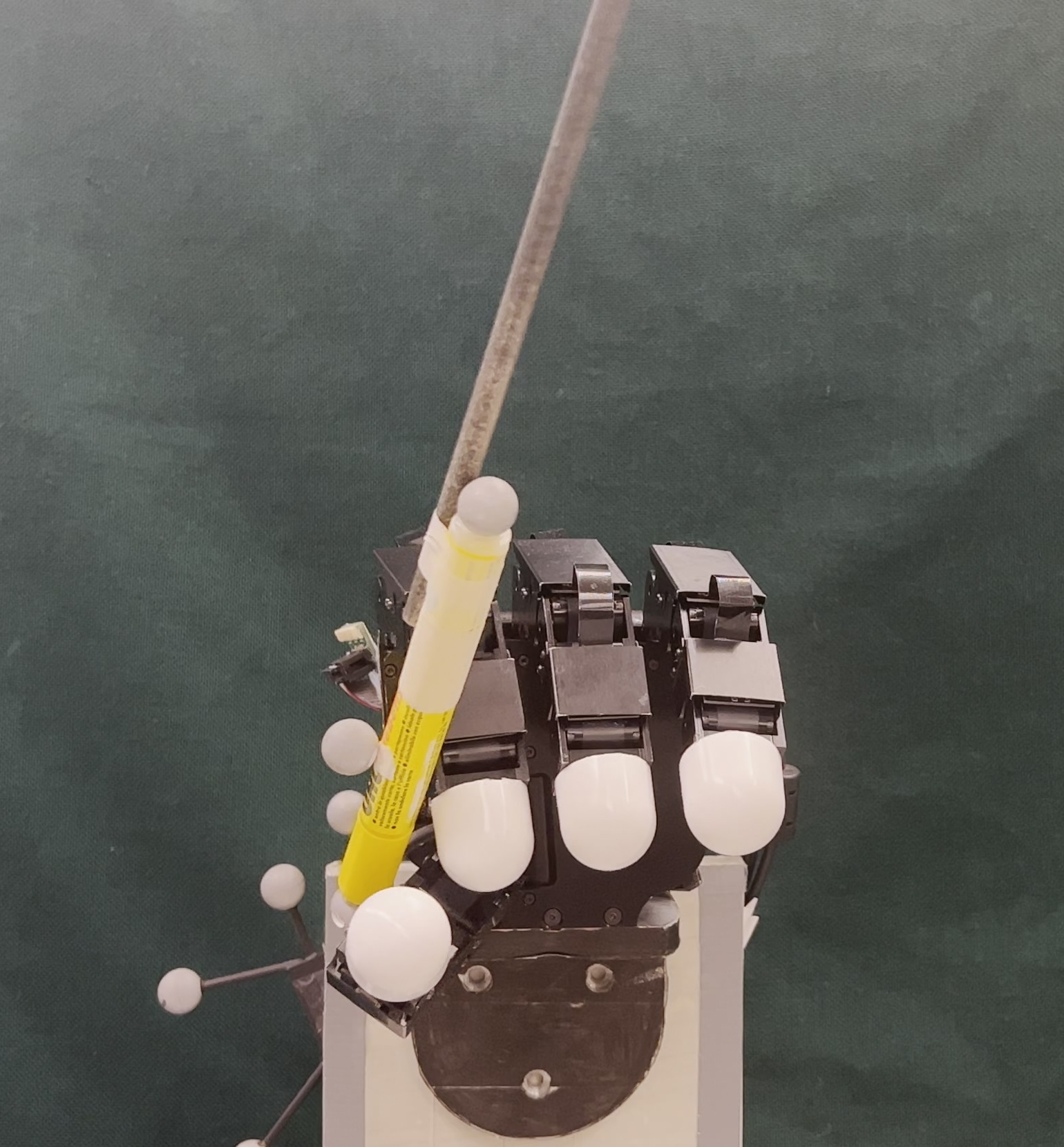}
            \caption{}
        \end{subfigure}  
        \begin{subfigure}[b]{0.15\textwidth}
            \includegraphics[width=\textwidth]{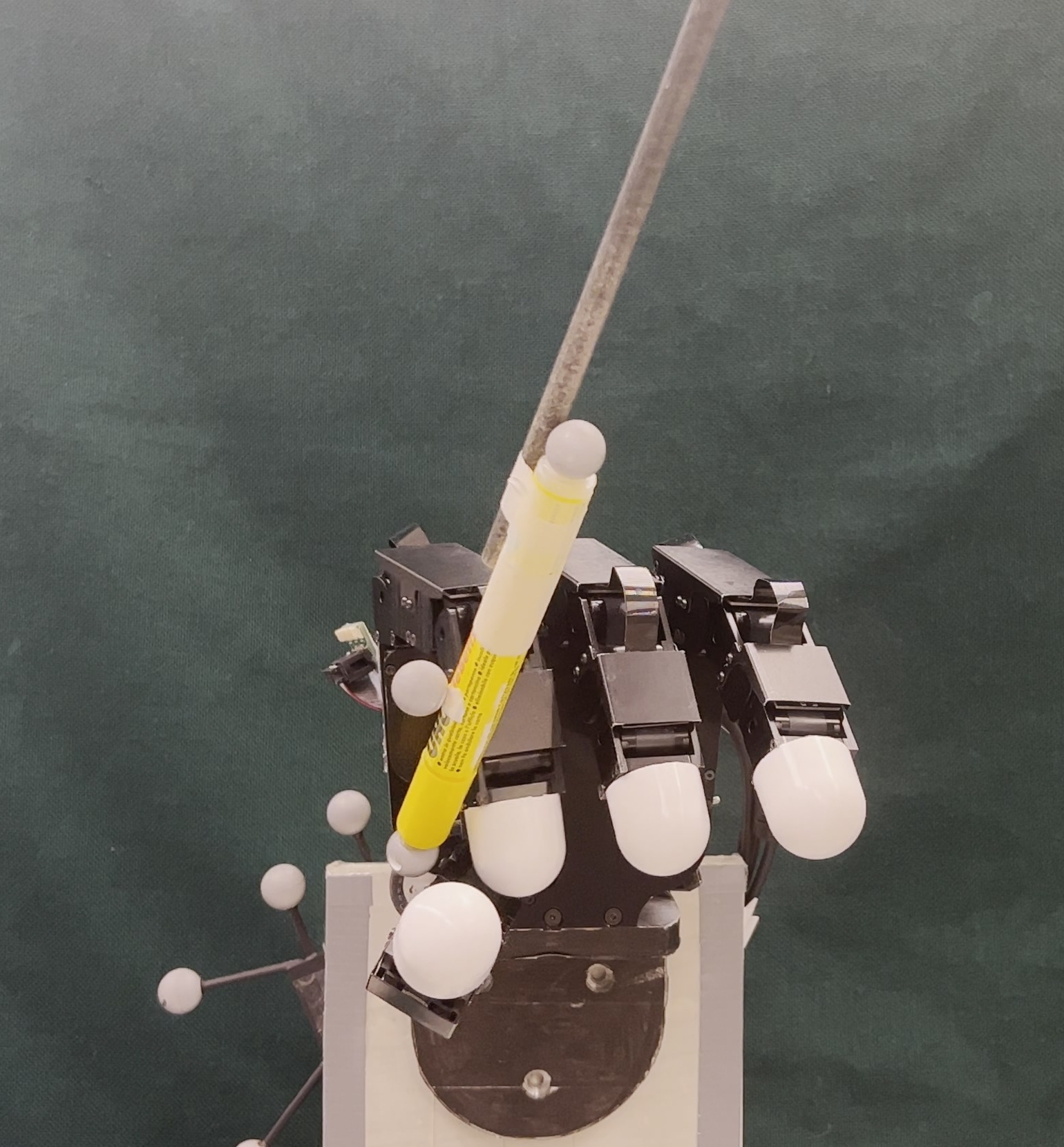}
            \caption{}
        \end{subfigure}
    \end{subfigure}
      \caption{Assessing the reactive collision avoidance approach on Allegro hand. Each row represents a trial. A glue stick (experimental object) approaches the hand. The hand is controlled to avoid both collisions with the object and inter-finger collisions. In scenarios (a) to (c), the object moves between two fingers, while in (d) to (f), it approaches the hand from the back towards the front and side of the fingers.}
    \label{fig:reactive_snapshots}
\end{figure}

\subsubsection{Collision-free motion planning for changing grasp pose}
The same experimental object was used in the second scenario. The goal is to change the grasp pose from a thumb-index pinch grasp (see Fig.~\ref{fig:experiment-stick}(1)) to a four-finger pinch grasp (see Fig.~\ref{fig:experiment-stick}(4)) by changing the placement of fingers.
We assume that the grasp remains stable during all intermediate states, allowing the object to be held without being dropped.
To generate a collision-free path, both DS-guided RRT* and dynamic PRM* were adopted.
This task is challenging because the ring finger needs to move from above to below the object within a very limited space inside the hand (Fig.~\ref{fig:experiment-stick}).
The thumb and the index finger were controlled to remain in the same joint configuration.
We only generated collision-free trajectories for both the middle and the ring fingers, leading to an 8D online motion planning problem.
Two different solutions were generated to achieve the task goal. In one solution (Fig.~\ref{fig:experiment-stick}(1)-(4)), the ring finger moves to the right side of the held glue stick to avoid potential collisions. An alternative solution is illustrated in the first row of Fig.~\ref{fig:regrasp_example}, where the ring finger moves left and then bends to avoid a collision; the middle finger also moves left to make space for the ring finger. Finally, both fingers arrived at the desired configuration.

\begin{figure}[t]
    \centering
    \begin{subfigure}[b]{0.48\textwidth}
        \includegraphics[width=\textwidth]{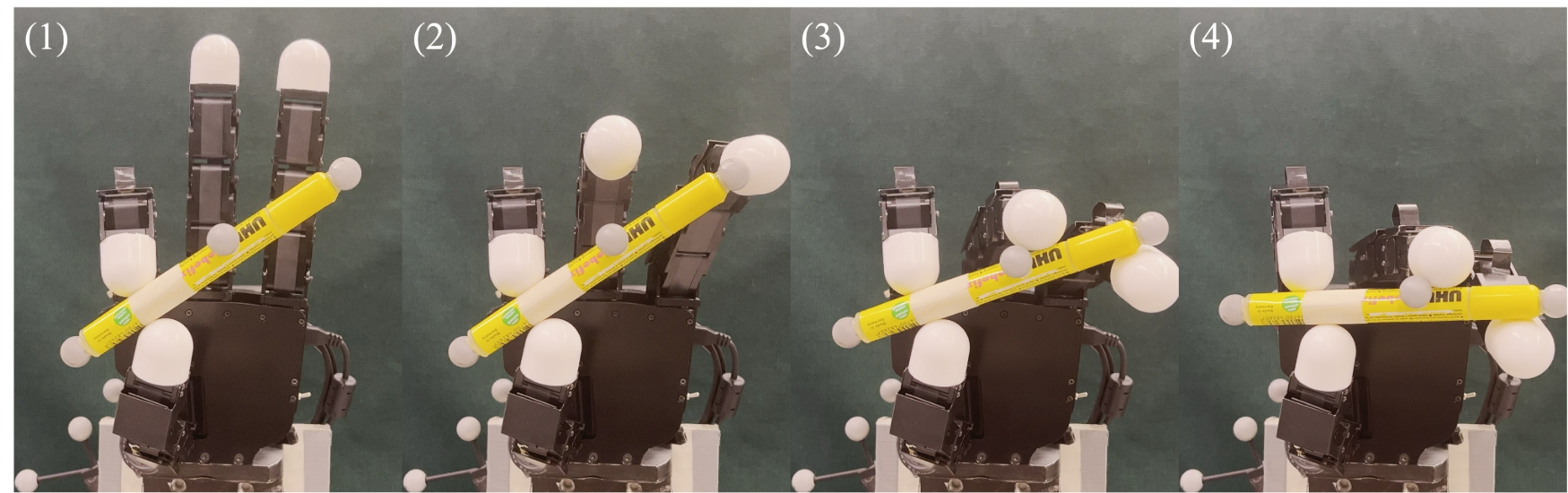}
        \caption{Collision-free finger replacement example: the ring finger moves beneath the stick to grasp it, avoiding collision, while the thumb and index finger pinch the stick, altering its pose through the movement.}
        \label{fig:experiment-stick}
    \end{subfigure}
    \begin{subfigure}[b]{0.48\textwidth}
    \includegraphics[width=\textwidth]{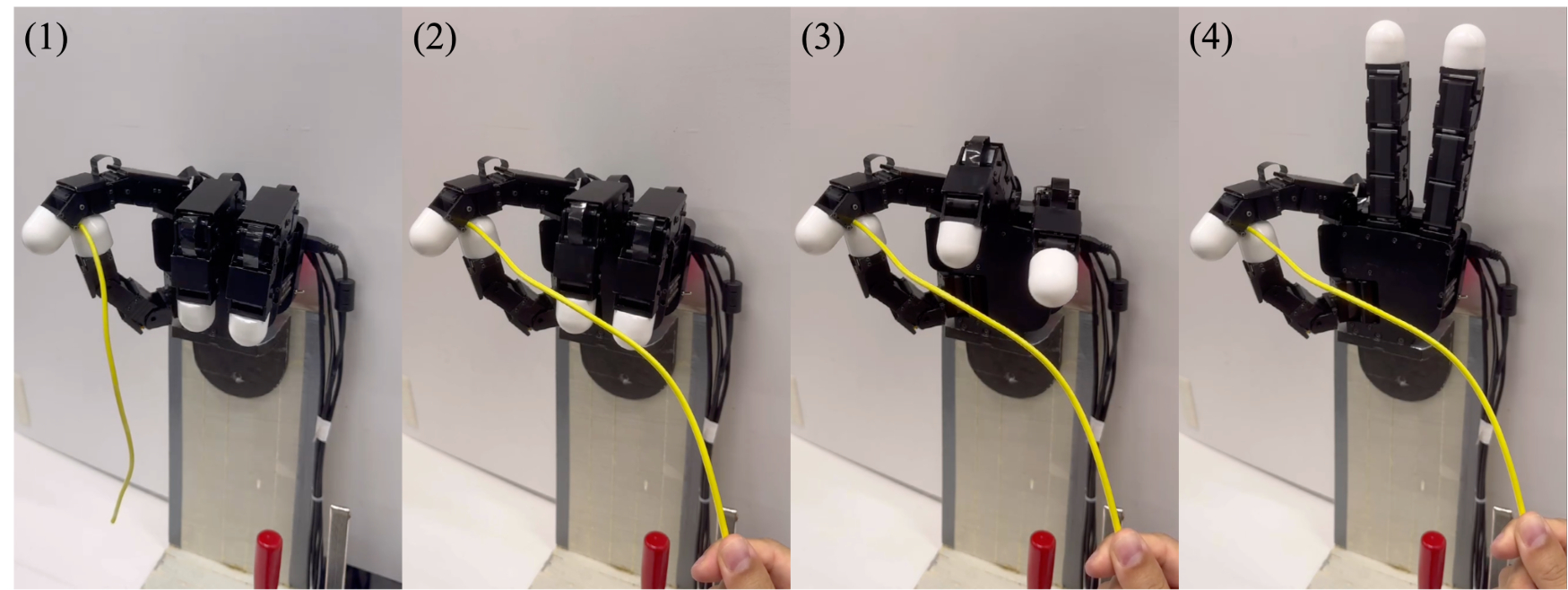}
        \caption{Example of online collision avoidance with a deformable object: the middle and the ring fingers avoid collision with a moving rope in real time during extension movement.}
        \label{fig:experiment-rope}
    \end{subfigure}
    \caption{Experiments of collision-free finger trajectory planning in real time.}
    \label{fig:regrasp_example_1}
\end{figure}

\subsubsection{Real-time collision avoidance with a deformable object}
In the third scenario, we use a deformable object to demonstrate that our algorithm does not require offline construction of object models.
We used a deformable yellow rope as our experimental object.
This rope is 0.36m long, with a 4mm diameter. The camera was employed to track the point cloud of the rope at 15 Hz. Images were segmented in HSV color space and then skeletonized. We use 50 points to represent the rope.
The robotic hand was mounted in the same configuration as the previous scenarios with the thumb and the index fingers pinching one tail of the rope with fingertips.
The task was to stretch the middle and ring fingers from bent to extended configuration, while avoiding collision with the rope, which was held and moved by the human experimenter (see Fig.~\ref{fig:experiment-rope}).
Experimental results demonstrate that our algorithms re-generated the finger trajectories in real time, according to the configuration of the rope, enabling the hand to successfully avoided collision with the rope.
 
\subsubsection{In-hand sliding manipulation}

The last scenario focuses on the regrasp with a four-finger pinch grasp, where the objective is to manipulate an object by making one or more fingers slide on its surface. We selected three objects from the YCB object set for this experiment: a banana, a fork, and a bottle. 
The same camera were adopted to track the poses of these objects in real-time.
Each object is represented by a point cloud of 200 points from a simplification of its mesh file. The point cloud was dynamically updated in real-time based on the object's tracked pose, and this updated data was then fed into the NNs. Utilizing the methodology outlined in Section~\ref{sec::sliding}, we generated sliding motions for the fingers (see Fig.~\ref{fig:sliding_exp}).

\begin{figure}[t]
    \centering
     \vspace{-0.3cm}
     \begin{subfigure}[b]{0.5\textwidth}
          \centering
          \begin{subfigure}[b]{0.3\textwidth}
           \includegraphics[width=\textwidth]{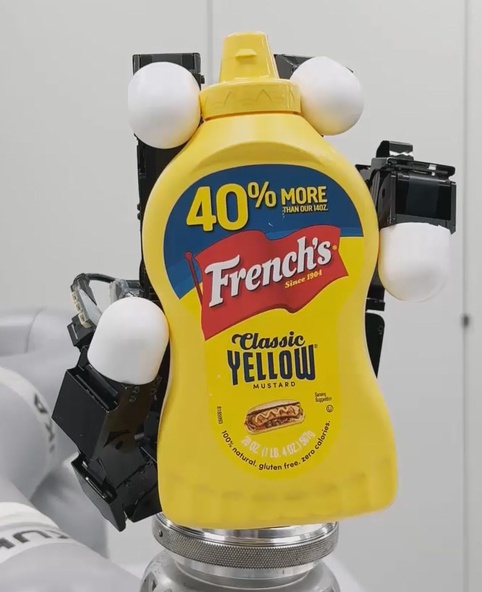}
            \caption*{}
        \end{subfigure}
          \hspace{-0.1cm}
          \begin{subfigure}[b]{0.3\textwidth}
          \includegraphics[width=\textwidth]{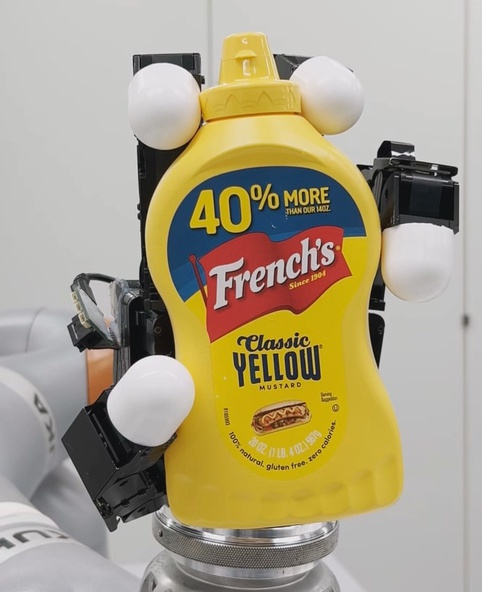}
            \caption*{}
        \end{subfigure}        
        \hspace{-0.1cm}
          \begin{subfigure}[b]{0.3\textwidth}
           \includegraphics[width=\textwidth]{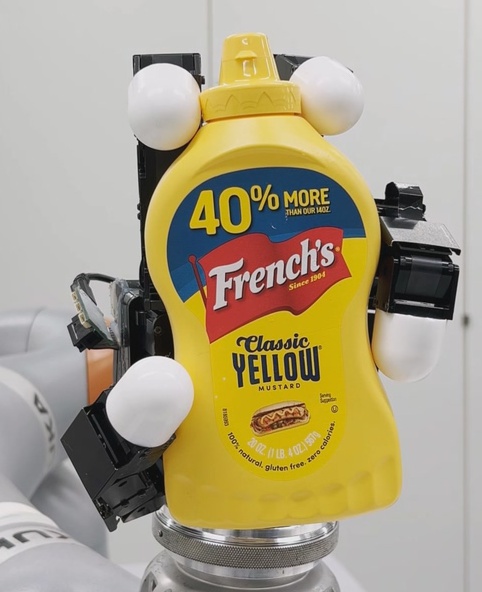}
            \caption*{}
        \end{subfigure}

       \end{subfigure}
       
        \vspace{-0.3cm}        
    \begin{subfigure}[b]{0.5\textwidth}
          \centering
          \begin{subfigure}[b]{0.3\textwidth}
           \includegraphics[width=\textwidth]{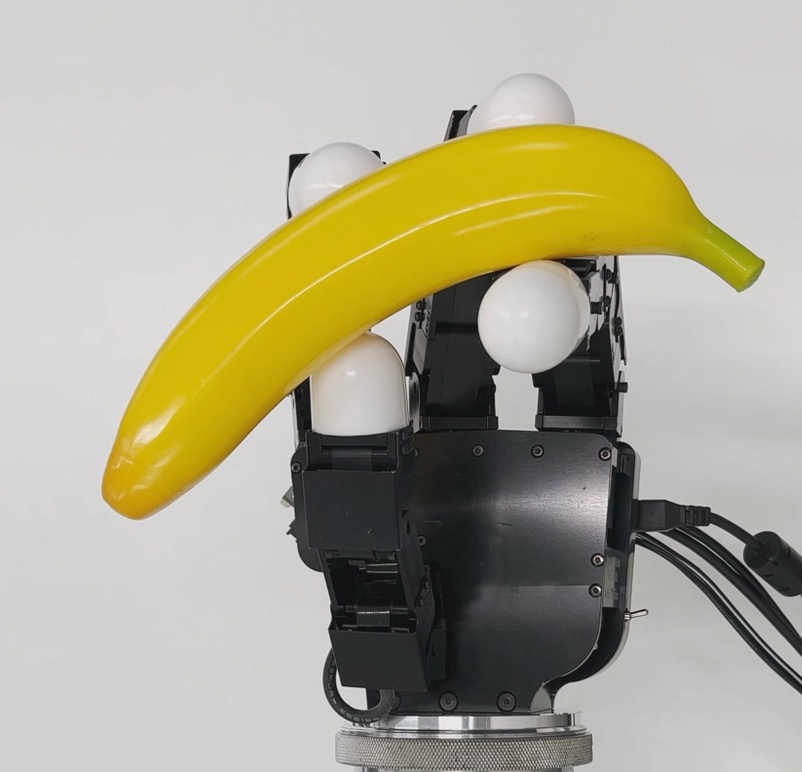}
            \caption*{}
        \end{subfigure}
          \hspace{-0.1cm}
          \begin{subfigure}[b]{0.3\textwidth}
          \includegraphics[width=\textwidth]{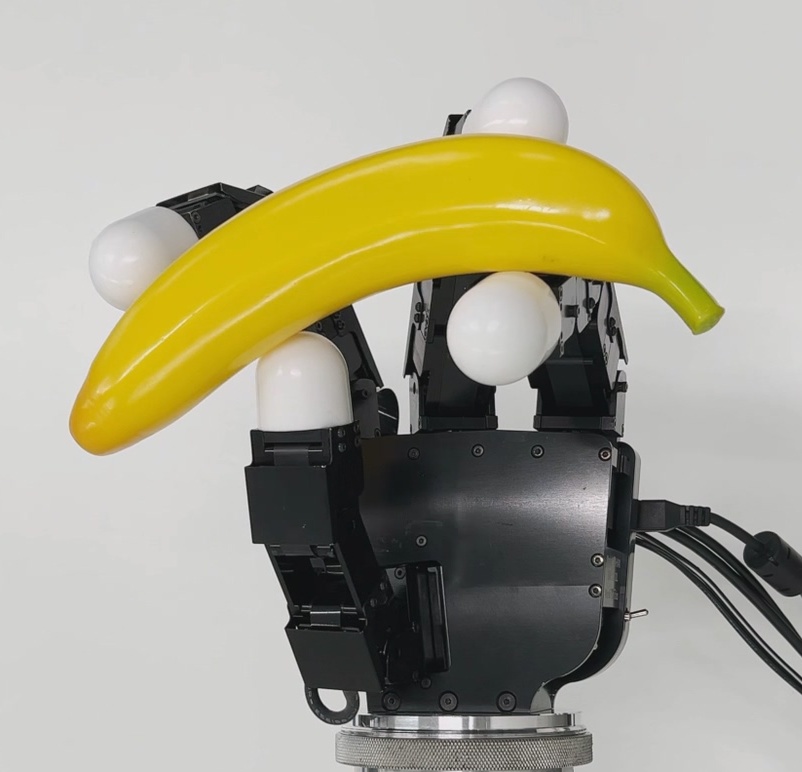}
            \caption*{}
        \end{subfigure}        
        \hspace{-0.1cm}
          \begin{subfigure}[b]{0.3\textwidth}
           \includegraphics[width=\textwidth]{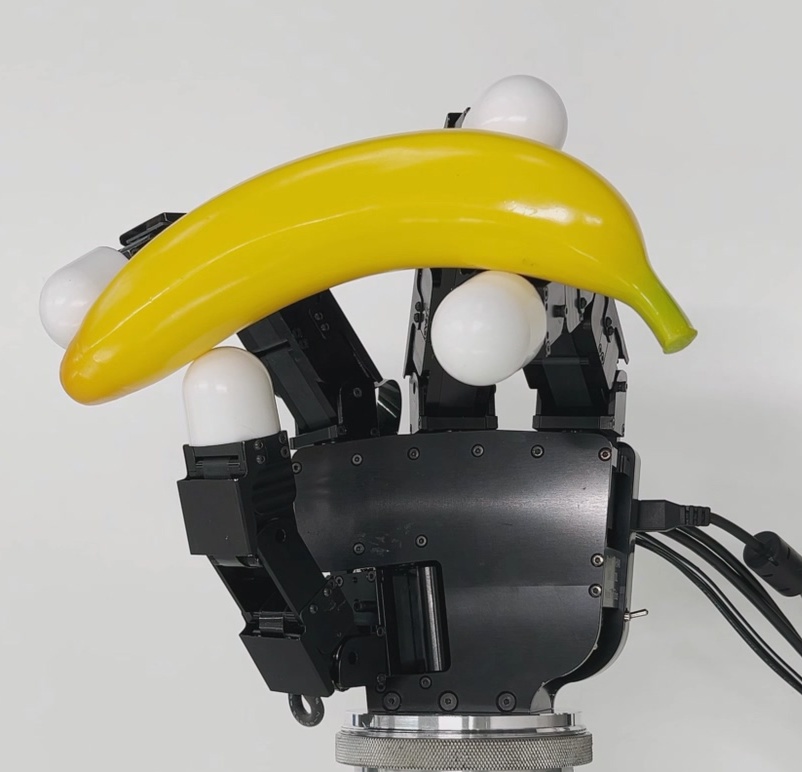}
            \caption*{}
        \end{subfigure}
       \end{subfigure}

      \vspace{-0.3cm}
     \begin{subfigure}[b]{0.5\textwidth}
          \centering
          \begin{subfigure}[b]{0.3\textwidth}
           \includegraphics[width=\textwidth]{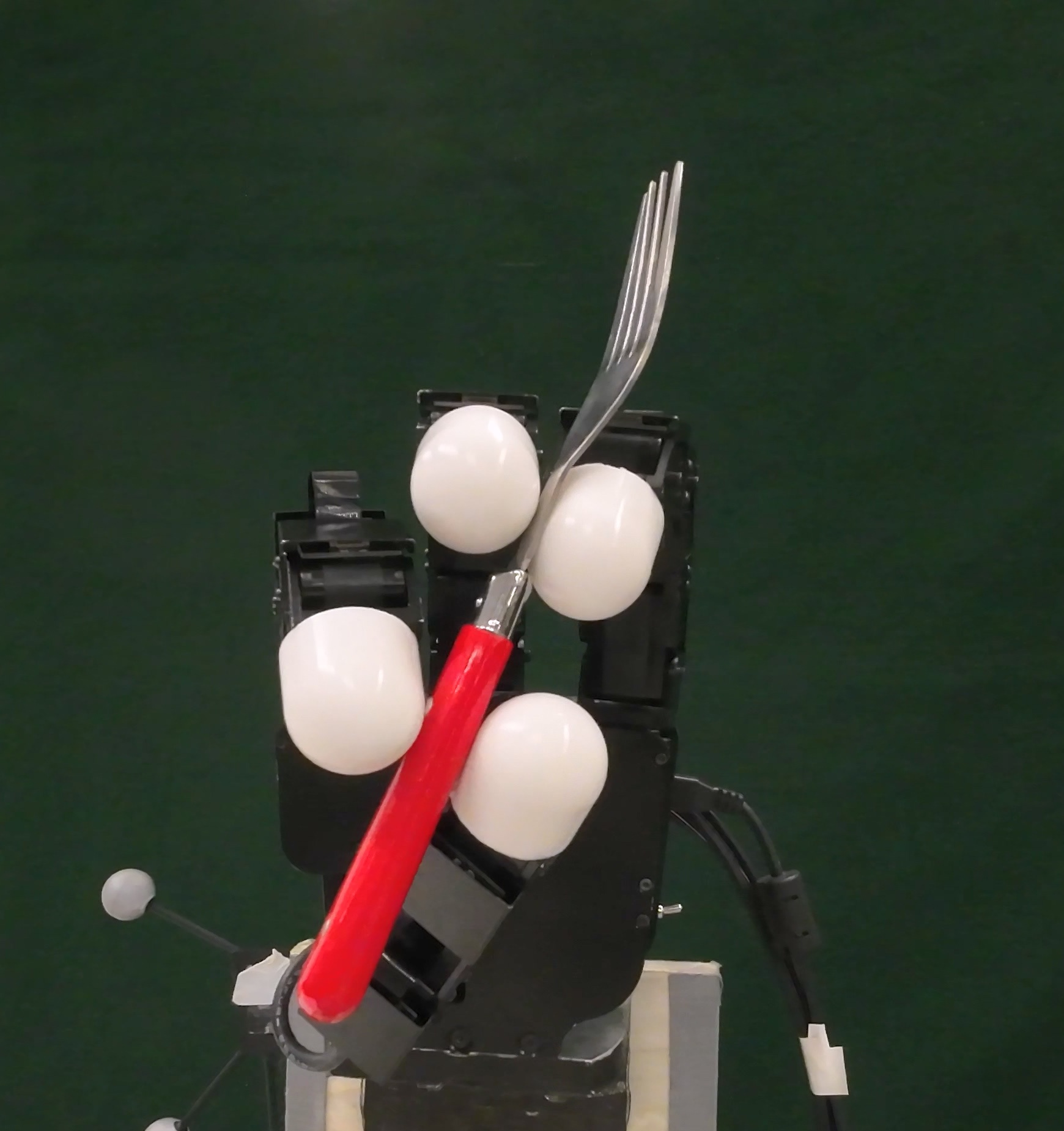}
            \caption*{}
        \end{subfigure}
          \hspace{-0.1cm}
          \begin{subfigure}[b]{0.3\textwidth}
           \includegraphics[width=\textwidth]{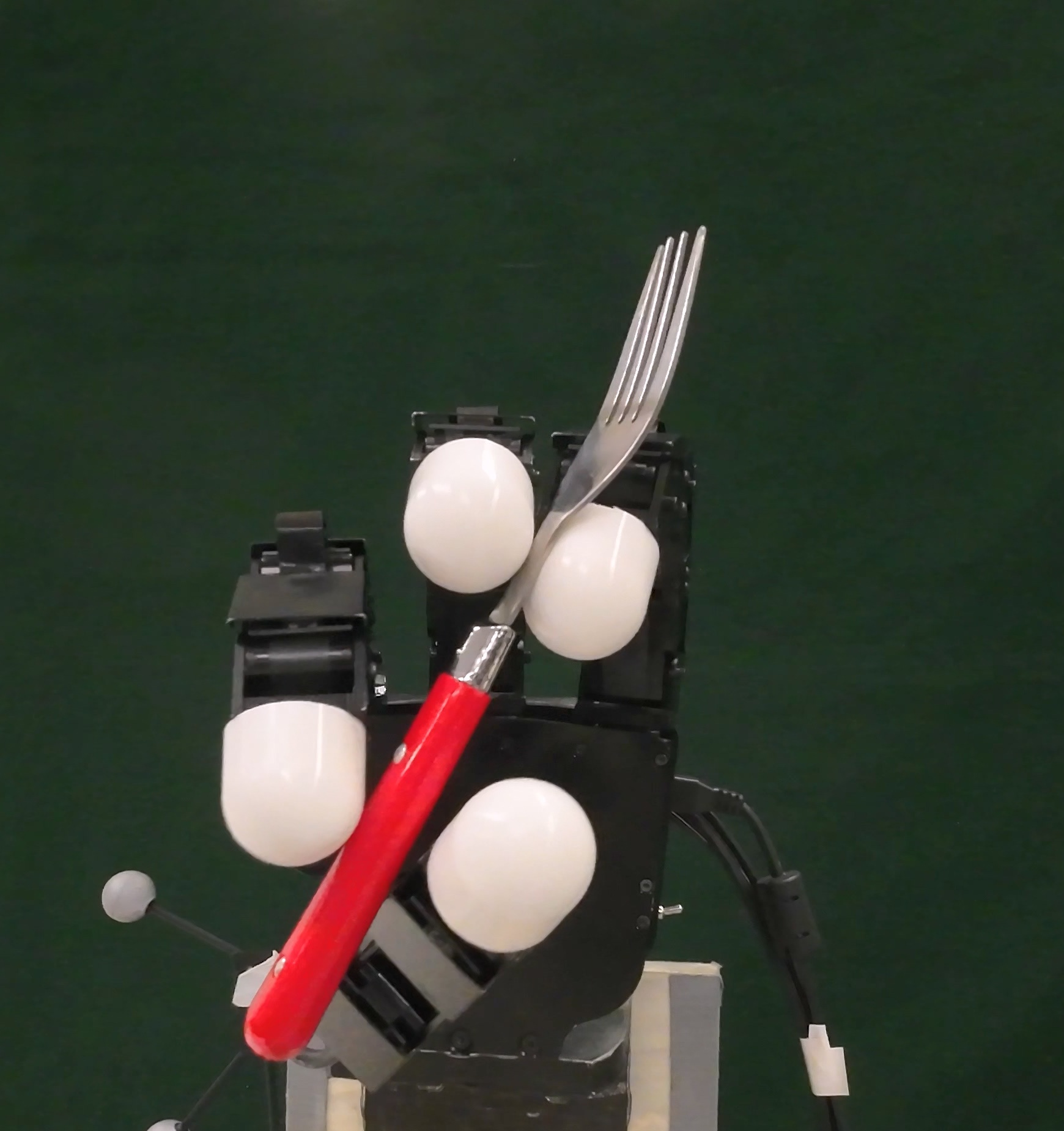}
            \caption*{}
        \end{subfigure}        
        \hspace{-0.1cm}
          \begin{subfigure}[b]{0.3\textwidth}
           \includegraphics[width=\textwidth]{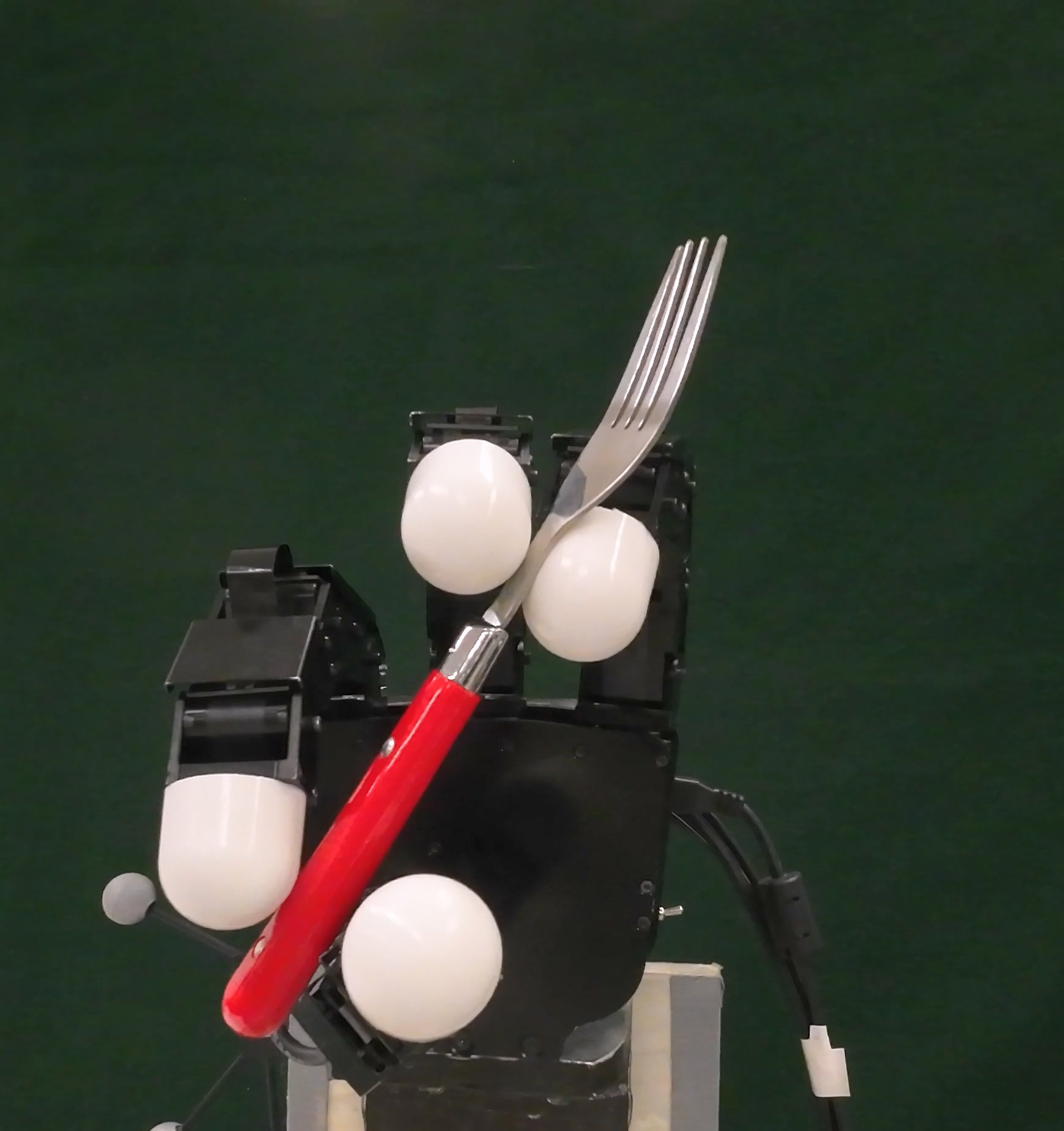}
            \caption*{}
        \end{subfigure}
       \end{subfigure}
    \caption{Examples for in-hand sliding manipulation. Our algorithm generates trajectories for the robotic fingers to slide on the object surface towards new grasping configuration.}
    \label{fig:sliding_exp}
\end{figure}

\section{Discussion and Conclusion}\label{sec::discussion}
In this study, we tackle real-time path planning challenges for multi-fingered robotic hands during in-hand manipulation, emphasizing grasp pose changes. We propose an NN-based approach to accurately represent C-space and detect collisions in real-time in cluttered, dynamic environments.
Our method utilizes the estimated distances and their gradient in C-space to guide a sampling-based motion-planning, RTT*, via DS, enhancing success rates in high-dimension problems.
Additionally, we have developed a dynamic PRM* approach that leverages parallel computation to enhance performance. Experimental results demonstrate that our approaches effectively address real-time collision-free path planning for a multi-fingered robotic hand.
Our model's applications go beyond collision avoidance, providing a precise object boundary and tangent depiction.
This representation potentially enables the planning of finger trajectories for a sliding motion on the object's surface or impedance control during manipulation given a specific penetration distance.
We plan to broaden our approach to various robotic systems and objects, investigating its scope for tasks like coordinated finger motion for dexterous in-hand manipulation.

Our paper presented two robotics motion planning algorithms, DS-guided RRT* and dynamic PRM*. While DS-guided RRT* can be implemented in real-time and improve the success rate in high dimensions, it may not result in the shortest possible path due to time constraints.
Conversely, dynamic PRM* improves the performance of PRM* but may still fail at finding a path in narrow passages that arise frequently in highly cluttered environments, a common limitation of PRM*.
One solution to this limitation could be to apply more colliding nodes as dynamic nodes in the PRM, so that these nodes are updated to collision-free nodes along the gradients and brought close to the collision boundary in C-space. Thus, more connected graphs could be built in narrow passages.
The lack of global information makes finding a collision-free path for optimization-based methods almost infeasible.
In a robotic implementation, failure cases may arise due to inaccuracies in localizing the object pose and tracking the robot's trajectory.
These limitations underline the need for further research in motion planning to address the current challenges and produce more efficient solutions.

\bibliographystyle{ieeetr}

\bibliography{refer}

\end{document}